# Comparing Spoken Languages using the Pāninian System of Sounds and Finite State Machines

Shreekanth M Prabhu[1] and Abhisek Midya[2]

[1] – Department of Computer Science and Engineering, Cambridge Institute of Technology, Bengaluru

[2] – Department of Computer Science and Engineering, Jain University, Bengaluru

## Abstract

The study of spoken languages comprises phonology, morphology, and grammar. The languages can be classified as root languages, inflectional languages, and stem languages. In addition, languages continually change over time and space by picking isoglosses, as speakers move from region to/through region. All these factors lead to the formation of vocabulary which has commonality/similarity across languages as well as distinct and subtle differences among them. Comparison of vocabularies across languages and detailed analysis has led to the hypothesis of language families. In particular, the Western linguists' view is that Vedic Sanskrit is a daughter language, part of the Indo-Iranian branch of the Indo-European Language family and Dravidian Languages belong to an entirely different family. These and such conclusions are reexamined in this paper. Based on our study and analysis we propose an Ecosystem Model for Linguistic Development with Sanskrit at the core, in place of the widely accepted family tree model. To that end, we leverage the Pāninian system of sounds to construct a phonetic map. Then we represent words across languages as state transitions on the phonetic map and construct corresponding Morphological Finite Automata (MFA) that accept groups of words. Irrespective of whether the contribution of this paper is significant or minuscule, it is an important step to challenge policy-driven research that has bedevilled this field.

**Keywords: Pānini's system of sounds, State Machines, Finite Automata, Phonology, Morphology, m-language, Phonetic Map, Comparative Linguistics, Ecosystem Model**

**Biographical Notes**

**Dr. Shreekanth M Prabhu** is currently working as Professor and Head of the Department of Computer Science and Engineering at Cambridge Institute of Technology, Bengaluru, India, His research interests include Social Networks, E-Governance, and Comparative Linguistics.

**Mr. Abhisek Midya** is currently working as an Assistant Professor in the Department of Computer Science and Engineering at Jain University, Bengaluru. His research interest is in Theoretical Computer Science.

Corresponding Author Email: shreekanthpm@gmail.com

Co-author Email: abhisekmidyacse@gmail.com

# 1. Introduction

Linguistics is a fascinating discipline going back millennia and has been a field for intense scholarly pursuit in India. Particularly among them are contributions by Pānini whose work on the system of sounds and formal grammar has inspired significant advances worldwide. Then there were generations of scholars enriching the field such as Kātyāyana, Patanjali, and Bhartṛhari. In recent times pioneering work by Chomsky has been the hallmark of the advances. According to Chomsky [1], the primary purpose of language is not communication, rather it is cognition as language is the primary vehicle for thoughts. Chomsky [2] also differentiated between I-language and E-language. Here I-language is a universal language that applies to all spoken/human languages. E-language caters to specific natural languages factoring in cultural and geographic aspects. Linguistics comprises phonology, morphology, and grammar. Broadly, phonology deals with the sounds in spoken languages; morphology pertains to the construction of words; and grammar describes the rules for the orderly usage of words to construct sentences.

By providing a structure to words and language, Linguistics makes understanding languages manageable. Otherwise understanding millions of words individually can prove to be daunting and time-consuming. Linguistics also helps in perpetuating languages. For example, thanks to a robust linguistic tradition of Sanskrit, ancient Ramāyana, written thousands of years ago is still intelligible to modern scholars. Without Linguistics, languages keep changing with time and place and literature becomes incomprehensible in a matter of a century or two.

The field of Linguistics has its roots in ancient India. The Vedas are preserved for millennia by oral transmission. To ensure accurate pronunciation, understanding, and appropriate usage of Vedic Hymns in Yajna, the scholarly tradition mandates the study of six Vedāngas as a pre-requisite and co-requisite for the study of Vedas. These six Vedāngas are Śiksha (phonetics, phonology, and pronunciation), Chandas (prosody), Vyākarana (grammar and linguistic analysis), Nirukta (etymology, explanation of words), Kalpa (ritual instructions), and Jyotish (astronomy). Here the first four have laid the foundation for Indian Linguistics. The expositions [3-6] give a very lucid explanation of ancient Indian Linguistics. In India knowledge is maintained using a 4-fold mechanism that includes Sutra, Vārtika, Bhāshya, and Kārika. Here Sutras are very compact, cryptic, and formulaic. Vārtikas are elaborations and Bhāshyas are interpretations of Sutras. Kārika captures the essence. Shyamsundar [7] has done an elaborate study of Pānini's contribution to linguistics and related it to the theoretical foundations of modern computing. Paul Kiparsky [8] elaborates on Pāninian Linguistics covering grammar, morphology, phonology, and phonetics,

There is a continuing grammatical tradition in India and Pānini's Astādhyāyi superseded all earlier traditions and core ideas from there spread to other languages and locales worldwide. Astādhyāyi draws on Vedic Sanskrit as it sets standards for classical Sanskrit. Patanjali's Bhāshya on Pānini's grammar is the most popular. The tradition has continued for centuries with newer Bhāshyas. Because of such rigorous discipline, the Vedas were transmitted without distortion for millennia.

Generally, linguistics can be approached from the viewpoint of words (Śabda) or sentences (Vākya). Whichever way you approach it both Śabda and Vākya are inextricably linked. The only purposeful way of using Śabda is in the form of Vākya. The only way to decipher and understand Vākya is by breaking it down into Śabdas. Vyakarana thus is called Shabda Śastra.

Pānini's Astādhyāyi analyses sentences; identifies words and then components, and arrives at Dhātus (roots of words). Each word is viewed as consisting of Prakriti (the original part) and Pratyaya (suffixes). By combining Prakriti and Pratyaya, the Padas (usable words) are formed. With a good discipline of grammar using a single Dhātu typically 360 words can be formed. There are at least 2000 Dhātus, resulting in lakhs of words. This framework enables Sanskrit to be a powerful language where new words can be easily composed using components and they become conveniently intelligible to those conversant with the language. When it comes to the right use of words, it can be done only with meaning in mind.

Three things are critical to interpreting the meaning of individual words in a sentence to arrive at the intended meaning of the sentence: Ākānkshā (expectancy), Yogyata(suitability), and Sannidhi(proximity). According to Vedic tradition, the six objectives of precise grammar are Rakshā (prevention from distortion), Asandeha (absence of ambiguity), Ūhā (modification of Vedic Mantras due to the possibility of more than one interpretation), Āgama (ease of augmentation), and Laghuh (easy means of acquiring knowledge).

In the last few centuries, Comparative Linguistics has emerged as a fertile field for fervid research. Here languages are compared for the similarity of words and then their structural properties. Using that approach linguistic families are formed, and even ancestral languages are hypothesized at times drawing far-reaching to far-fetched conclusions about the history of populations and their movements. Not just languages, but literary sources can also be considered sources of words.

The bulk of work related to comparing languages concerns itself with comparing words across them. Comparing the words also may mean comparing root words, inflections, and derivations. This generally calls for specialist know-how from the field of linguistics. In many cases, there are disputes as different linguists draw differing conclusions based on their perspectives.

According to modern Comparative Linguistics, certain words are considered isolates i.e. they are unique to a given language or a narrow set of languages. Then there are isoglosses. Isoglosses are picked from neighbouring languages or speakers, including those transiting through an area. The isoglosses cause dialectical variations from region to region. These differences due to isoglosses may be phonological, lexical (different words), or structural. Cognates sound similar across languages carrying the same/related meaning. The cognates are classified as adstrate words when these are loan words due to trade and migration. Then there are substrate words where it is presumed that speakers of one language dominated the speakers of other languages resulting in an asymmetric transfer of words. In addition, according to substrate theory, the native speakers who speak an intruding language also affect it phonetically, lexically, and structurally. In contrast, in Indian tradition, the words in a language are divided into three categories: Tatsama (same as words in another language generally Sanskrit), Tadbhava (derived from words in another language, typically Sanskrit), and Deshya (native words). Further, polyglossia is very typical in India, where different languages/dialects are used in differing contexts, in the same region, by the same people.

The research in comparative linguistics has been overly influenced by the worldview of researchers, leading to 'policy-driven research' where researchers resist outcomes not to their liking. As a result, significant findings of some scholars recede to the background and drop out of discourse. There is a strong need to address this lacuna to arrive at a balanced perspective.

In this paper, we primarily focus on morphology, and specifically how the words move across languages. Firstly, we leverage Pānini's system of sounds and construct a phonetic map where each sound has a unique coordinate and each word is a path/walk on the map. The distances between words can also be gauged on the map. Secondly, each word is represented as a sequence of state transitions, where each sound is a state and the succeeding sound is the input to guide the transition. Thirdly, for each word group, we propose an m-language (formal language that recognizes a set of valid words) that uses an m-alphabet (set of sounds) using Finite Automata. The word groups can be extended and inter-connected. Each m-language will have a core alphabet and an extended alphabet. Then, we make use of insights gained from our analysis to revisit the assumptions and conclusions of linguists and strive to arrive at an alternate model that is more robust. Herein we draw on ecosystem literature.

The rest of the paper is as follows. Section 2, *Literature review* surveys the scholarly work in Comparative Linguistics hitherto. Section 3, *State Machine Model for Comparative Linguistics* covers how we model sounds using a phonetic map based on Panini's system of sounds and then word formation within and across languages using Finite State Machines. Section 4, *Discussions*, revisits the antiquity of Vedic Sanskrit, language formation, and word formation and proposes an Ecosystem Model for Linguistic Development with Sanskrit at the core. In Section 5, *In Retrospect*, we go beyond scholarly linguistics and view the field with a holistic perspective. Here we cover significant studies of language enthusiasts who studied current Indian languages driven by curiosity and passion. This is followed by the contribution of scholars, and philosophers who immersed themselves in Indian traditions, cultures, and philosophy and then studied Indian languages. Section 6, *Our Contribution* contextualizes our work in fields such as Computational Morphology, Machine Learning, and Vector Databases, distinguishes it from conventional approaches taken hitherto, and highlights future research avenues. Section 7, *Vocabulary* describes the vocabulary we have compiled for this work. Section 8, *Conclusions* concludes the paper.

## 2. Literature Review

The relationship between languages did not get the attention of scholars in Europe as according to Biblical tradition, Hebrew was considered the universal language which then broke into other languages. In India, Sanskrit was considered the mother of all languages while scholars were very much aware of Sanskrit words and words native to a given language. In Europe, as acknowledged by Mallory [9], James Parsons [10] was probably one of the first to do a systematic study of thousands of common words across European Languages. However, according to Mallory [9], a century before that it was Joseph Scaliger who attempted to divide the languages of Europe into four major groups, each labelled after their word for God. The transparent relationship of what we today call the Romance languages was recognized in the 'Deus' group (for example, Latin 'Deus', Italian 'Dio', Spanish 'Dio', and French 'Dieu'), and contrasted with the Germanic 'Gott' (English God, Dutch God, Swedish 'Gudy' and so on); Greek 'Theos'; and Slavic Bog (such as Russian 'Bog', Polish 'Bog' and Czech 'Buh'). This exercise of comparing languages was also undertaken by visitors to India in the 15th century. In India, it was Filippo Sassetti and Thomas Stephens were the first two who noticed the similarity between Indian and European Languages. Singh B [11] identifies Thomas Stephens as the first Englishman in India. Pedro Redondo [12] explains that the motivation of Sassetti was that of the humanist whereas that of Stephens was evangelical and theological. All these exercises and the well-known discourse of William Jones [13] culminated in the proposal of not only the Indo-European Family of Languages but also the acceptance of the language family as a universal construct.

Initially, Sanskrit was considered the mother of the Indo-European Languages as it had cognates across Indo-European Languages and the most complete grammar with eight cases as well as duals in addition to singular and plurals. But then scholars who are generally known as Indologists who call themselves mainstream changed their stance. Bryant [14] puts forward the 'main-stream' view that (i) There has to be a proto-language probably spoken by all speakers before that broke into Indo-European (IE) Languages; (ii) All the IE speakers stayed in a common homeland before they separated; (iii) The proto-language could not have been Sanskrit; (iv) There was Proto-Indo-European(PIE) Language that broke into Celtic, Germanic, Romance, Baltic, Slavic, Greek, and Indo-Iranian families with PIE at the root. Thus, Sanskrit was relegated as a leaf node within the Indo-Iranian family, and India became yet another outpost of IE speakers.

Bryant explains how Sanskrit was dethroned using linguistic arguments. One of the reasons linguists proposed PIE is that Sanskrit has innovated a(pple), e, and o sounds to 'a(pathy)' sound, the first sound among Sanskrit vowels. Greek has retained the original sounds. A typical example given is that 'bhend' in Greek becomes bandh in Sanskrit. Another example the scholars give is Greek Deca (for number 10) is not derivable from Sanskrit Daśa, hence there needs to be a common ancestral language to both. The languages are further classified as Centum and Satem languages based on the word for the number 100 and here Kentum Languages are considered more archaic. Sanskrit is considered Satem Language and was ruled out as an archaic language and by extension any language from India. However, Burushaski, a language spoken in the Gilgit-Baltistan in India was found to be a Centum Language.

Further, since Sanskrit had retroflexes, which many European languages did not have, some linguists say it cannot be a protolanguage. To support their hypothesis scholars claimed that Sanskrit borrowed cerebralization from Dravidian Languages and any word in Sanskrit that is not in common with European Languages is a loan from Dravidian or Munda languages. This contrasts with Indian tradition where Sanskrit words appear either as Tatsama or Tadbhava forms across languages and seldom the other way around. As an example, the word for water is Neer only in Sanskrit and Dravidian Languages but not in most Indo-Aryan Languages. Hence, one may conclude that the word was loaned into Sanskrit. However, any such conclusion may be hasty as Greeks use neró for water, which is likely from Sanskrit.

Witzell [15] is a prominent proponent of the mainstream view that Aryans are outsiders to India and that the Vedic language is an import into India and import of Munda words into Vedic Sanskrit, presumably after Vedic Language came into India. These words are not found in other Indo-European Languages. Kuiper [16] considered such loan words (and features) to be of Dravidian origin. Their conclusion of Sanskrit being a branch of Indo-Iranian are not borne by other studies. For instance, Dr. Gintaras Songaila [17] elaborates on enormous affinities that are directly there between Indo-Aryan and Lithuanian without any connection with the Iranian language.

Bryant and Patton [18] examine the issue of Indo-European origins from multiple perspectives in an edited volume. Among the linguists who contributed to that endeavour, Mishra [19] claims that Sanskrit is more archaic than all others. The main features where Sanskrit is shown to deviate from Indo-European is the merger of IE 'a', 'e', and 'o' into 'a' in Sanskrit and the change of palatal k, etc. to palatal s, etc. in Sanskrit. Mishra counters this and among many other arguments gives the example of the Gypsy language where Indo-Aryan 'a' remains '*a'* in Asiatic Gypsy but becomes *a*, *e*, *o* in European Gypsy. This confirms that the original IE *a* was the same as Sanskrit *a* and remained *a* in the Indo-Iranian languages, but changed to *a*, *e*, *o* in their sister languages. The distortion of 'a(pathy)' sound in Sanskrit continues to play out even in modern times as Americans pronounce Kamala Harris as Camela Harris. Dr Geoff Lindsay [20], a British Linguist delves into the topic of vowels used by British and Americans, how they have a hard time pronouncing the vowel 'a', the first Sanskrit vowel, and how they tend to use a(pple).

Mishra gives the case where Sanskrit retains both Vākya and Vāchya. According to Mishra, ś becomes k before it becomes s in Sanskrit. He maintains that ś and k are allophonic. Thus, the k which was allophonic to ś in Sanskrit might have been generalized in the Centum languages. He also gives examples of Lithuanian, a Satem Language sporadically presenting the 'k' sound. Mishra also points out that Schleicher [21], Bopp [22], and well-known linguist Grimm originally accepted that Sanskrit 'a' is the original IE vowel and Greek 'a', 'e', and 'o' as later development. It was only later that scholars reversed their stance. This was based on their understanding of palatalization, which Mishra challenges.

Linguists also noticed common words between Uralic and Indo-European Language families and speculated that this happened as Indo-Europeans were heading towards India. Mishra cites Harmatta [23] who did a detailed study and identified large number of common words which were transferred to Uralic Languages. Hermatta considered them as Indo-Iranian words. Mishra identifies them as Indo-Aryan words. Hermatta classified these common words belonging to 11 different periods, the oldest being 5000 BCE and the youngest being 1500 BCE. All these transfers are in one direction. There are no Uralic words in Indo-Aryan. Mishra considers this as evidence of Indo-Aryan languages continually migrating through the Ural region over centuries. Hermatta referred to them as Indo-Iranian words, which on closer analysis Mishra found to be Sanskrit words. Table 1 below lists the Sanskrit words and periods of transfer.

Table 1: Sanskrit words transferred to Uralic Languages

| Period | Sanskrit word: meaning | Approximate Period Given by Harmatta |
|---|---|---|
| 1 | aj:drive | 5000 BCE |
| 2 | arbhaka: child, bhagah: god, mrtah: dead, daivah: heaven | 4600 BCE |
| 3 | ashTrah:whip. chaagah:goat | 4200 BCE |
| 4 | argah:gift given to guest | 3800 BCE |

| 5 | dhenuh:cow, dadhi;milk (related), svasa;sister | 3400 BCE |
|---|---|---|
| 6 | vrshah:bull, sapta;seven, dasha:ten, shata:hundred, rashmi; strap, cord | 3000 BCE |
| 7 | maksi: honey bee, madhu; honey, yavah:corn | 2700 BCE |
| 8 | asurah: lord, sarah; flood, sura: beer, wine, Sahasra: thousand | 2400 BCE |
| 9 | shosah; to become dry, aksharah: booklet, rill | 2100 BCE |
| 10 | Visham: anger, hatred, ara:bowl | 1800 BCE |
| 11 | ankh:hook | 1500 BCE |

Mishra also studied words in a Hittite source related to horse training. He identifies the Sanskrit words in Hittite. The period associated with these findings is between 1600-1500 BCE. Mishra identifies characteristics in Hittite that are common with Middle-Indo-Aryan. He also disputes Laryngeal theory, based on which linguistics consider Hittite archaic. Table 2 below lists, Sanskrit words found in horse training manual with their equivalent.

Table 2: Sanskrit words in Mittani Horse Training Manual.

| Sr No | Sanskrit word | Word in the Training Manual | Meaning |
|---|---|---|---|
| 1 | vasanasya | washannnashaya | stadium |
| 2 | rathya | aratiyanni | part of cart |
| 3 | ashvani | asuvanni | stable master |
| 4 | babhru | babrunnu | Red brown |
| 5 | bharita | baritannu | Golden yellow |
| 6 | pingala | pinkarannu | Red yellow, pale |
| 7 | rukma | urukammannu | jewel |
| 8 | jira | jirannu | quick |
| 9 | magha | makanni | gift |
| 10 | marya | maryannu | Young warrior |
| 11 | mati | matunni | wisdom |

Based on his analysis, Mishra asserts that the language found in horse training manuals is conclusively Indo-Aryan. It is neither Iranian nor Indo-Iranian.

Subhash Kak [24] makes a long list of common words among European languages and Sanskrit. He emphasizes the contiguity of central Asia with India from ancient times. The borrowing of words also spans disciplines, 'Astipathi' in Sanskrit becomes osteopathy, and 'Jara' the word for old age in Sanskrit leads to geriatrics. The same is true with the common medical word sputum which has a natural association with Sphut, a Sanskrit word than spit, an English/Latin verb. The English word 'pāth' is due to 'path' in Sanskrit (as used in Rajpath i.e. King's Road) leading to words such as allopathy and homeopathy. Hence the transmission of words has continued for centuries and millennia.

Table 3 below has a list of Sanskrit words and cognates contributed by Subhash Kak. Herein, we refer set of sounds used in a word as m-alphabet (morphological alphabet).

Table 3: Sanskrit Words and Cognates in European Languages

| **Sanskrit Word** | **m-alphabet (Sanskrit)** | **Word (Language)** | **m-alphabet (Extended)** | **Sounds Gained** | **Sounds Changed** | **Sounds Lost** | **Related words/meaning** |
|---|---|---|---|---|---|---|---|
| āvāsa | a,ā,v,s | house (En)<br>haus (Ge) | a,ā,v,s,h,o,u | h, o, u | v to u | | |
| dam | d,a,m | Dom (Ru)<br>Damus (La) | d,a,m,o,u,s,h | o, u, s ,h | a to o | | domicile, domestic |
| grha | g,r,h,a | Casa (La)<br>Cass (Sp) | g,r,h,a,k,s | k, s | g to k, s to h | r | home |
| vāri | v,r,ā,i | Water (Du) | v,r,ā,i,t | t | | i | |
| udaka | u,d,k,a | Uda (Ko)<br>Voda (Ru) | u,d,k,a,v | v | u to v | | water |
| āp | a,p | Apa (Ro) | a,p, | | | | water |
| nīr | n,r, ī | Nero (Gr)<br>Dur (We)<br>Neeru (Ka) | n,r, ī, d, u | d, u, o | n to d | | water |
| dhara | dh,r,ā | Terra (It)<br>Dal (We) | dh,r,ā,d,l,t,e | d, l, t, e | dh to t, dh to d | | earth |
| nabha | n,a,bh | Nebo (Ru, Cr)<br>Nebe (Cz) | n,a,bh,b,e,o | b, e, o | bh,a | | sky |
| Varuna | v, r, ṇ, u,a | Ouranos (Gr) | v, r, ṇ, u, a,o | o | v | | |
| yuva | y,u,v,a | Youth (En)<br>Jeunesse (Fr) | y, u, v, a, | t, h | | | Juvenile |

Thus, the formation of cognate words may involve sound shifts, and closely related sounds (voiced versus voiceless, aspirated versus unaspirated, changes of vowels) as well as changes to grammar (gender-related or other changes) or due to any other peculiarities of receiving languages. Thus, we can define a grammar that can cater to such scenarios which can determine if a word belongs to a word group or not. Additionally, we may be able to generate candidate words that can prospectively belong to the same word group.

Also, few studies compare Dravidian Languages with other Indian languages. A study by Swaminath Aiyar [25] is a rare exception. Aiyar after a unique and highly detailed comparative study of languages says "My views differ from those of all previous scholars because they contended themselves with comparing Dravidian Languages with Classical Sanskrit and naturally saw no deep-seated affinities. When one language is extensively affected by another, we need to look for the source of influence not in the artificial language of high literature but in the spoken idioms of common people. It is necessary to compare Dravidian idioms with the Vedic Dialects and the Prākrits of pre-Christian Centuries before we can decide the question of Aryo-Dravidian affinities". It was Bishop Caldwell who compared Classical Sanskrit and Dravidian Languages and pronounced the differences. At the same time, there were other scholars such as Pope, who also was a missionary did not agree. Pope opined the decision to consider Dravidian Languages as disjoint from Aryan Languages was rather abrupt. He expressed the opinion "(i) that between the languages of Southern India and those of the Aryan family, there are many deeply seated and radical affinities and (ii) that the differences between the Dravidian Tongues and Aryan are not so great as between the Celtic (for instance) languages and the Sanskrit; and (iii) that by consequence the doctrine that the place of Dravidian dialects is rather with the Aryan than with Turanian families is still capable of defence". He illustrated these positions using copious illustrations and pointed out that "the resemblances appeared in the most uncultivated Dravidian dialects' and that "the identity was most striking in the names of instruments, places, and acts connected with a simple life". He promised to follow on with a paper that looked at derivative words and showed that the prefixes and affixes were Aryan. The work of Aiyar thus fills that gap.

The Drāvidian Languages were historically divided into the Andhra Group with Telugu and a set of languages and the Dravida group consisting of Tamil, Kannada, Malayaḷam, and Tuḷu. Andhra Group is independently influenced by neighbouring Prākrats as well as a greater propensity to use Sanskrit words. Aiyar's main conclusion is that in addition to many clear Sanskrit (Tatsama) words in the Drāvidian Languages, a significant number of Tadbhava words are derived from Sanskrit. He claims that when Caldwell came up with the hypothesis that Dravidian Languages have a low affinity for other Indian Languages, he compared words from Classical Sanskrit which indeed were different for the sample he had chosen. Aiyar invalidates Caldwell's conclusions by comparing South Indian Language words with other Sanskrit words that are closer to Vedic Sanskrit, Prākrits, and other Indian Languages. Table 4 contrasts Caldwell's approach with that of Aiyar's.

Table 4 Comparison of Sanskrit and Tamil Words

| Sr,No. | English Word | Sanskrit Word (Caldwell) | Tamil, Telugu, Kannada, Malayalam | Proposed Word (Aiyar) | Remarks |
|---|---|---|---|---|---|
| 1 | hair | kesha | Mayir(Ta) | Śmashru(Sa) | |
| 2 | mouth | mukha | Vay(Ta) | Vac(Sa) | Vac is alternate word from Vedic Sanskrit |
| 2(a) | nose | | Mūkku(Ta), Mūgu(K), Mukku(Te) | | Words derived from Mukha are used for face and mouth. Here it is proposed to be used for nose as well |
| 3 | ear | karna | Shevi(Ta) | Śrava(Sa), shravika(Sa) | |
| 4 | hear | sru | Kel(Ta) | Karna(Sa) | |
| 5 | eat | bhaks | Tin(Ta) | Trṇu(Sa), Tr(Sa), | |
| 6 | walk | car, cel | Egu(Ta) | Ya(Sa), i(Sa) | |
| 7 | night | nak | Ira, Iravu | Rātri(Sa) | |
| 8 | mother | matr | Āyi(Ta) | Yāyi(Paisc.) | |
| 9 | tiger | vyaghra | Puli(Ta) | Vengai(Tamil) | |
| 10 | deer, beast | mrga | Marai, Man, Ma(Ta) | Mrga(S), Maga(Pr_ | |
| 11 | Fire | Agni | Ti(ta) | Tejas(Sa), Tij(Sa) | |
| 12 | Snake | Sarpa | Pāmbu.(Ta), Aravu (Ta), Arava(Ma) | Prasarpa, Sarpa, Sarpaks | |
| 13 | Village | grama | Ūr(Ta), Ūru(Ka) | Pura(Sa) | |
| 14 | buffalo | mahiSa | Erumai(Ta), Emme(Ka) | Heramba(ka) | Associated words are swapped |
| 14(a) | | | M āDu(Ta) | MahiSa(Sa) | |
| 15 | horse | ashva | Kuthirai(Ta) | Ashvatara(ka) | |
| 16 | hill | parvata | Malai(Ta) | Paruppu(Tam) | Matching Associations found |

According to Swaminath Aiyar, a large number of Dravidian words, in particular in Tamil that appear to have no affinity with Sanskrit are Tadbhava words from Sanskrit. As Tamil has a highly constrained Alphabet(sounds), they went through a lot more transformation and corruption compared to North Indian Vernaculars and appear unrelated. To get the whole picture one needs to look at a plurality of Sanskrit words and Prākrit words and inter-relationships between Dravidian Languages, as the closest word could belong to Telugu or Tamil in most cases and then further transformed in modern Kannada and Malayalam. Table 5, contains a sample of words analyzed by Aiyar and inferred as Sanskrit words. Aiyar derives Dravidian words from Sanskrit/Prākrat words with a variety of rules such as sound elision, sound substitution, and suffix additions.

Table 5: Tadbhava Dravidian Words which are derived from Sanskrit

| Sr. No | Sanskrit Word | Meaning | Tamil/Dravidian Word/Other Indian Language | Meaning |
|---|---|---|---|---|
| 1 | Paksha | Wing, Side | Pakka(Ta) | Side |
| 2 | Pashya | See, Look | Paar(Ta), Paḷe(Ko) | See |
| 3 | Dakshina | South | Tenkaṇa(Ta) | South |
| 4 | Bhru | Brow | Pubbu(Ta), Hubbu(Ka) | Eyebrow |
| 5 | Satya | Truth | Sari(Ka), Sahi(Hi) | Correct |
| 6 | Vayalah | Bangle | Baḷe(Ka), Vaḷai(Ta) | Bangle |
| 7 | Lokah | People, Word | Olaku(Ta) | People, World |
| 8 | Mridu | Soft | Mella(Ka) | Slowly, Gently |
| 9 | Mrda | Mud | Maṇṇu(Ka),Maṇṇ (Ta) | Soil, Earth |
| 10 | Dhvani | Voice, Sound | Toni(Ta) | Sound |
| 11 | Vandyah | Barren Woman | Banje(Ka), Vandi(Ta) | Barren woman |
| 12 | Shabdah | Word | Sadd(Pu), Saddu(Ka) | Sound |
| 13 | kāṣṭakah | Wood | Koṭṭai(Ta), Kaṭṭige(Ka) | Wood (Collected from Forest) |
| 14 | Mrtya | Perishable (Body) | Mai(Ka) | Body |
| 15 | Svithra | Silver/White | Velli(Ta), Belli(Ka), Belagu(Ka). Belaku | Silver, White,Light |
| 16 | Sreṇi | Line | Eṇi(Ka) | Ladder |
| 17 | Chayah | Hand | Kai(Ka, Ta) | Hand |
| 18 | Śirah | Head | Sir(Hi), Tale(Ka), Tare(Tu) | Head |
| 19 | Kārṣapaṇa | Coin or weight | Kāṇam(Ta) Kāhavaṇo(Pr) Kāhāṇ(Or) | |
| 20 | Meṣa | Sheep/Goat | Meḍam(Ta), Meke(Ka) | Goat |

According to Aiyar, the original Dravidian Languages were under the influence of Aryan Languages from the early days. He claimed after omitting clear Sanskrit words, there may be 1000 root words in Dravidian Languages. The tense and mood signs are highly influenced by Indo-Aryan Languages. In conclusion, he says the basic portion of Dravidian vocabulary consists largely of words of Indo-European origin. But owing to the extremely limited character of Tamil and Dravidian Alphabet (sounds), these words have been greatly corrupted and are very difficult to recognize as similar. In addition, he identifies around a hundred suffixes in Dravidian languages used for indicating tenses and modes of verb forms as of Aryan origin.

He disputes the contention of other scholars that Dravidian Languages have influenced Vedic Sanskrit. He claims cerebralization of sounds in Sanskrit is internal development. Dravidian Languages all along have retained a few alveolar forms from historic times and two still retain them. They have no particular preference for cerebral sounds via-s-vis alveolar sounds or dental sounds. Languages like Telugu do not tolerate cerebral sounds ṣ and ṇ. Other changes in Indian Languages are due to the transition from the synthetic stage to the analytical stage. In summary, he says Dravidian scholars have mistaken the reflection for the original and the original for reflection.

The words analyzed by Aiyar [25] are reproduced in Annexure 2[26]. Aiyar demonstrates that numerous common-place words in Dravidian Languages are Tadbhava forms from Sanskrit. One only needs to trace the transformation journey.

## 2.1 Research Opportunity

The worldview of Europeans is guided by the prism of conflict and conquest, leading to theories such as invasion theory and substratum theory. India indeed was subject to conquests from the 7th century AD onwards which targeted Indian civilization with religious conversions and political conquests. However, the essential characteristics of the civilization that survived have been convergence, confluence, continuity, and contiguity aided by amalgamation, and assimilation. Thus, India has a continuing civilization going back millennia and a sense of unity that stems from identification with the larger sacred geography unified by common traditions, scriptures, belief systems, holy places, and value systems. Diana Eck [27] rightly observes that India is a country united by the footsteps of pilgrims. The migrations of people within India have been continuous, across classes. Migrating priestly classes have maintained essential unity of traditions. Many southern kings also have northern lineages. Such movements have resulted in far greater homogenization of languages across India. The languages which were neighbours to the Sarasvati River region such as Konkani and Punjabi are inflectional like Vedic Sanskrit. The South Indian Languages tend to have more agglutination of consonants and less conjunction of consonants. However, subject-object-verb order is common across all Indian Languages.

Further, the larger geography which included Afghanistan and Central Asia was considered contiguous to India with cultural transmission and exchange. The Central Asian Republics continue to use 'Sthan'(place) as part of their names (Kazakhstan, Tajikistan) showing the influence of Sanskrit on them. Greater India thus consisted of Uttara Kuru as well as Uttara Madra regions. Another point to be considered is the Sinhala language of Sri Lanka located to the south of Dravida region is Indo-Aryan with commonality with Vedic Sanskrit retaining a few rather archaic words.

Sanskrit for most of the time served as the lingua franca across India thus serving as the donor. language of words that represented abstract concepts on one hand to mundane reality on the other. In Sanskrit, refined and accurate pronunciation was not only important for rituals but also considered a hallmark of the civilized. Generally, Apabramsha(mispronounced) forms of a Sanskrit word that is easier to pronounce were used by the commoners. Thus, Śrāvan word for the rainy season changed to Sāvan in Hindi. We notice that some languages (Kannada, Konkani, Bengali) retain the original. The word for cotton Karpasa is considered to have derived from Kāpas a Munda word. But other Indian Languages (Konkani, Marathi, and Gujarati) use Kāpas only, they are far away from Munda-speaking regions. Some argue that Kāpas is Apabramsha for Karpasa and not necessarily a loan word from Munda. In India, the direction of changes is from Sanskrit to Prākrit to vernaculars as India had a tradition of Chandas (language for prosody) and Bhasha (language for common use) concurrently evolving. This runs counter to the linguists' view where they expect the transformation to happen from simple/primitive to refined.

In addition, different regions of India and languages there have shown a preference for certain sounds and a lack of preference for others. Thus, the retroflex sound 'ṇ' is not in vogue in Hindi, but very much there in Konkani, Marathi, and Punjabi. Bengali uses o instead of a and 'b' sounds instead of 'v', in certain cases. In Bihar, 's' sound is used more than the 'ś' sound used in western regions. In contrast, Bengal which is located to the east of Bihar uses ś sound in place of 's' sound. On the other extreme, Iranian languages have replaced 's' with 'h'. In many cases Sanskrit has more than one sound, say for people Jana is used as well as Gaṇa is used. The same is true with Dik and Disha both words are used for direction in Sanskrit but for different cases. Further, Sanskrit uses a word starting with K for Kendra (center) which very few European Languages (Greek, Armenian), use, and most use centrum which starts with the 's' sound.

Thus, analysis of European Linguists using their worldview and rules may need revisiting using a formal approach that can address voluminous vocabulary across languages. In particular, Sanskrit commonly has more than ten alternative words to represent the same entity or concept. European Languages are generally compared only with Sanskrit, but not as much with other Indian Languages. It is also worth comparing the phenomena that Indian Language words underwent as they carried forward Sanskrit words and comparing the same with what could have happened to Sanskrit words that are borrowed by/found in common with European Languages. Figure 1 illustrates the transfer and transformation of words in Indian Languages. Here most spoken languages are derived from Prākrit and then further embellished by words from Classical Sanskrit. In the case of languages such as Konkani spoken close to the Sarasvati River, certain distinctive archaic features of Vedic Sanskrit are retained that are not found in Marathi despite both being very similar [28].

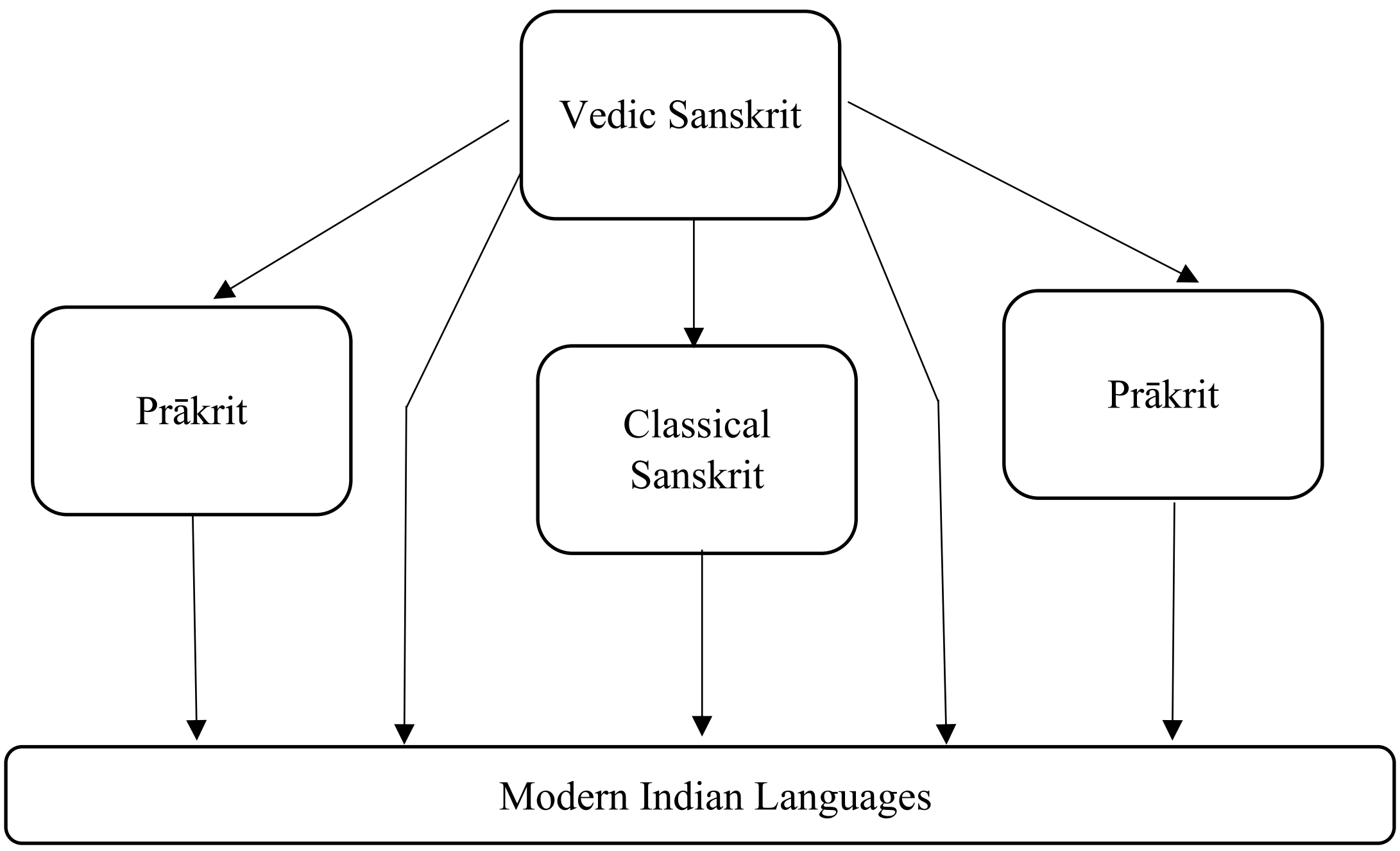

Figure 1: Word transfer and transformation in Indian Languages

The above generic representation could apply more broadly, beyond Indo-Aryan Languages. Based on an analysis of linguistic structure, Mishra [19] claims that Greek is closer to Middle-Indo Aryan and Dravidian is akin to the early stage of Indo-Aryan. Unfortunately, Mishra [19] and Aiyar [25] were deceased before their work could get published. Their findings have not got due attention, in particular affinity of Dravidian Languages with Vedic Sanskrit.

In summary, the dethroning of Sanskrit as a protolanguage and positioning her as a sibling of Greek needs to be revisited. In the least, confining Sanskrit as a daughter language under the Indo-Iranian branch is a travesty. Further, the inter-relationship between Dravidian Languages and Indo-Aryan Languages needs many more studies.

# 3. State Machine Model for Comparative Linguistics

Modern linguistics like ancient linguistics comprises phonology (the science of sounds), morphology (word formation using sounds), and grammar (deriving new words and constructing sentences). Analyzing the sentences thus consists of syntax analysis, semantic analysis, and pragmatics. The methodology for the analysis of natural language can be compared with the approach taken by the compiler to analyze programming languages. A compilation process consists of a scanning phase where a statement is broken into components (lexemes) and then in the parsing phase, a syntax tree is constructed comprising of lexemes and validated for grammatical correctness. Even though natural language processing is similar, the grammar is not context-free and morphology (the constructions of words) itself makes use of grammar in addition to the construction and analysis of sentences. However, some key constructs such as finite automata and the concept of language from theoretical computer science can be leveraged. That is the endeavour of this paper.

In this section, we introduce the concept of m-alphabet which is the set of phonemes used to construct a word. The core m-alphabet is the set of sounds that pertain to the original part (Prakriti) of the word, that too where the chosen sounds are common cutting across languages or that pertain to the suspected original word. The m-languages consist of words belonging to a word group that are related phonetically, semantically, grammatically, and ontologically. The word groups across different languages are compared and analyzed using these morphology-based constructs. We make use of Pānini's System of Sounds which represents natural language sounds comprehensively in a scientific manner.

## 3.1 Pānini's System of Sounds

Pānini developed the system of human/natural language sounds after a careful study of how they are generated by the vocal box. Pānini's Śikṣa (phonology) explains the form of each Varṇa ((letter/sound) is determined by Svara (intonation), Kāla (time taken to pronounce it), Sthāna (place of articulation), and Karaṇa. Abhyantara Prayatna (effort within the oral cavity) and Bāhya Prayatna (effort outside the oral cavity) are two additional factors. Figure 2, illustrates Pānini's System of Sounds.

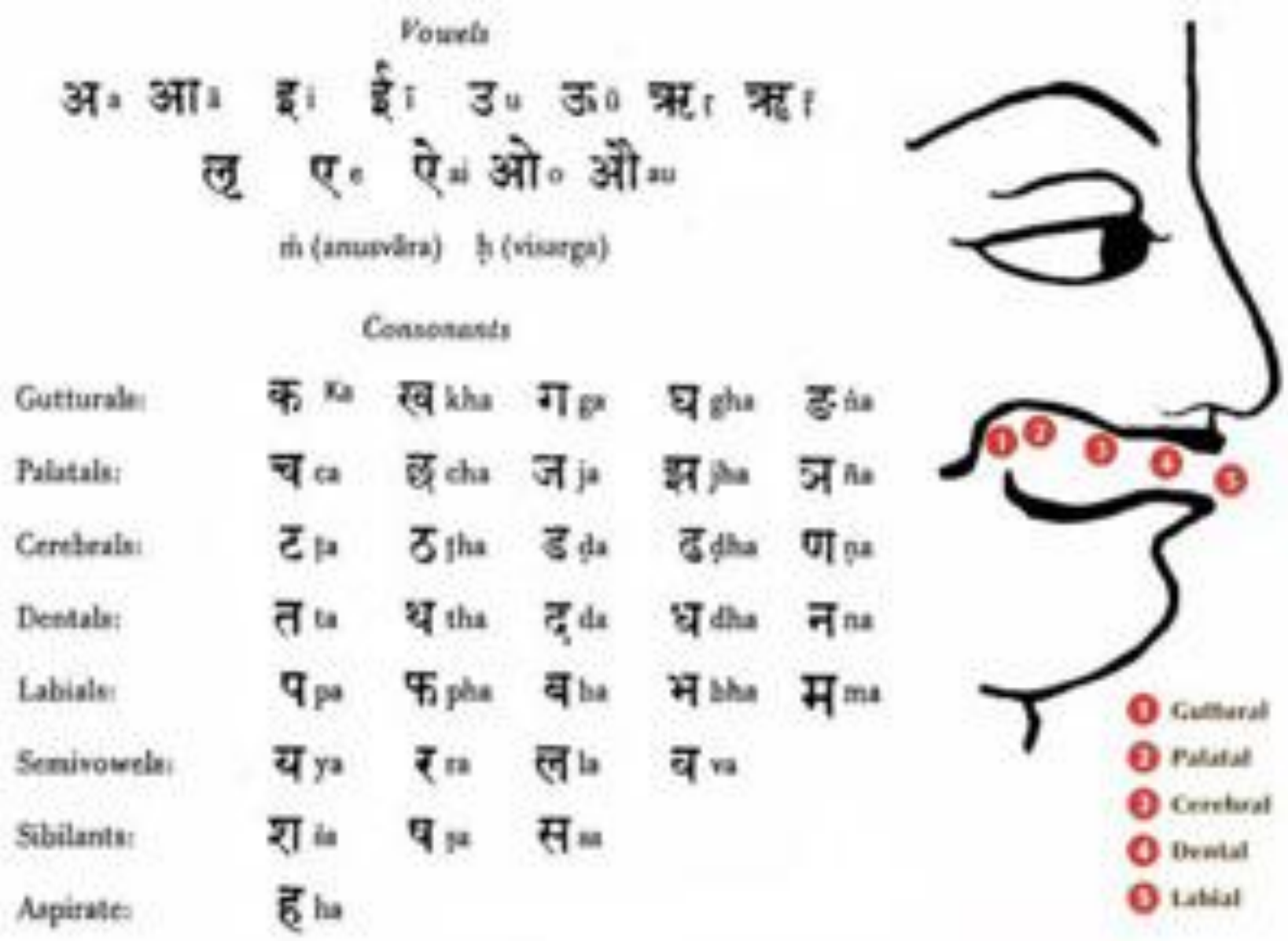

Figure 2: Pānini’s System of Sounds

Sounds that do not face any obstruction when we speak are termed vowels. These may vary depending on whether they are short, long, or very long. In his scheme, there are 13 vowels and two additional vowels which can be used only in conjunction with other sounds namely am and ah. The sounds that face obstruction are termed consonants. He classifies them based on place of articulation. The guttural/velar/Kanṭavya sounds are produced in the throat. Next, palatal/ Tālavya Sounds are generated by touching one’s tongue to the pallet. The next set of sounds are Cerebral/Murdhya sounds. They are also called hard palatal sounds or retroflex sounds as it requires one to reverse the direction of the tongue while generating them. The fourth set of consonants is dental/Dantavya. They are generated by touching the tongue to the teeth. The fifth set of consonants is labial/Austa. Here the lips are involved in generating the sounds. Each of these groups of 5 consonants can be further classified – (i) unvoiced and unaspirated/tenuis ii) aspirated, (iii) voiced (iv) voiced and aspirated, and (v) nasal.

Then there are other consonants which are called semivowels, sibilants, and aspirates. Figure 1 below illustrates Pānini’s System of Sounds. Rajesh Kumar [29] and Anuradha Chaudhari [30] explain Pānini’s system of sounds covering modern linguistics and traditional Indian vocabulary.

Whereas Pānini’s System of Sounds is very comprehensive and representative, some sounds are not represented specifically. Vedic Sanskrit and many Indian Languages have a cerebral ḷ sound which is at times used instead of the ḍ sound as in Iḍa, and Iḷa. This is not represented above.

Alveolar sounds are intermediate sounds typically used when English say “Tea”, “Table” or “Tennis”. They are not fully dental. A person who is a native speaker of a language that has retroflex sounds; may treat them as such. Then there are additional alveolar sounds in Tamil which are not there in North Indian Languages. Tamil and probably other Dravidian Languages early on had far too limited an alphabet or far fewer phonemes. Tamil continues to have a limited alphabet consisting of vowels: a, ā, i, ī, u, ū., e, ai, o, ō, au, with the omission of r, rr, lr. The consonants are k, nasal (k), c, nasal(c), t, n, ṭ, ṇ p, m, y, r, l, v, *l, l, r, n.* The last four are alveolar sounds and are unknown to Sanskrit Alphabet. In each class of consonants, instead of 5 members, only tenuis (the first), and nasal (the last) sounds are there.

Generally, European Languages do not use cerebral/retroflex sounds, except in a few North European Languages such as Swedish. Some languages such as French use only dental sounds. The Tamil Language also has far fewer sounds and the script uses the same symbol for four consonants of the same category.

Further, there are a total of nine fricative consonants in English: /f, θ, s, ʃ, v, ð, z, ʒ, h/, and eight of them (all except for/h/) are produced by partially obstructing the airflow through the oral cavity. These are: /f/: far, /v/: save, of, /θ/: think, /ð/: those, /s/: sir, race, /z/: zoo, rise, /ʃ/: sharp, chef, pressure, sugar, motion, /h/: ahead.

## 3.2 Analyzing Words using Sounds

In this section, we build a word bank cutting across languages. Table 6 indicates the encoding we have used for the languages.

Table 6: Encoding to indicate the language of the word

| European Languages | Indian Languages |
|---|---|
| English (En), German (Ge), Russian (Ru), Greek (Gr), Romanian (Ro), Latin (La), Latvian (Latv), French (Fr), Lithuanian (Li), Italian (It), Welsh (We), Danish (Da), Dutch (Du), Spanish (Sp), Polish (Po), Portuguese (Por), Bulgarian (Bu), Corsican (Co), Croatian (Cr), Uranian (Uk), Scot Gaelic(SG), Irish (Ir), Slovak (Sl) | Sanskrit (Sa), Prākrit (Pr), Hindi(Hi), Marathi(Mar), Punjabi(Pu), Konkani(Ko), Bengali(Be), Gujarati, Kannada(Ka), Tamil(Ta), Telugu (Te), Malayalam(Ma), Sinhala(Si) |

Subhash Kak did a study of words derived from Sanskrit in European Languages. Table 7 below lists Sanskrit words, and corresponding cognates in European Languages. We have also added a word in Kannada and Konkani for water.

Here we also list basic sounds used in Sanskrit words which we call m-alphabet (Morphological Alphabet). This is followed by an extended alphabet to represent all words, sounds gained, replaced, and lost. Also listed are related words. All words in a given row can be considered to constitute an m-language (Morphological Language).

The words from Vedic Sanskrit have gone through a variety of transformations in Indian Languages. This is accepted by all. Now we hypothesize that the transformation of those words in European Languages can also be considered the manifestations of the same phenomena that happened as the words got carried over to European Languages. For example, Graha in Sanskrit becomes Kar in spoken Punjabi but in Hindi, it remains as Ghar. Thus, it is not just European languages that use the 'k' sound.

Tables 7 to 11 illustrate the concept of m-alphabet and m-language with additional examples which we have collected. Note that this is based on Google Translate output and our knowledge which may have missed certain synonyms that are cognate. Annexure 1[31] has a bank of Indian and European words, which we have enumerated. nearly two hundred groups of words for which m-languages can be defined.

Table 7: m-language for word group "Being in the middle"

| Theme | Being in the middle, in between |
| --- | --- |
| m-language | madhya (Sa), mādhyam(Sa), middle, medium, mediate, media , midten(Da), midden(Du), madhala(Ma), madhyama(Ka), milieu(Fr), mezzo(It), mitte(Ge), meio(Po), mijloc(Ro), maeda(Si), meadhan(SG), mesaio(Gr)} |
| Non-members | natuttara(Ta), lar(Ir), vidu(latv), vidurio(Li),sredina(Ru) |
| m-alphabet(core) | {m, d, y, a,i} |
| m-alphabet (Extended) | {m, d,y, a, I, t, n, l, c, z} |
| Remarks | Sanskrit, Indic, Germanic, Greek and Romance language and Scot Gaelic, use the above m-alphabet. |
| Extended Vocabulary | mezzanine floor, meso (between micro and marco) |

Table 8: m-language for word group "Face, Mouth"

| Theme | Face, Mouth |
| --- | --- |
| m-language | mukh(Sat), moga (Ka)}, muh(Hi)}, mouth, mukhya(Sa:Main), mund(Da), mond(Du), mute(Latv), tond(Ko) |
| Non-members | Face, Chehera(Hindi), beul(Irish), Bayi(Kannada) Usta(Slovenian) |
| m-alphabet(core) | {m, u, kh,o,g, t, n, h,d} |
| m-alphabet (Extended) | {m u, k, kh, h, o, g, y, d, n, t} |
| Remarks | Face and mouth words get overlapped. Tond may belong to another m-language with Sanskrit Connection, Tunda – trunk. Germanic and Sanskrit languages have commonality. |

Table 9: m-language for word group "Long. Tall"

| Theme | Long, Tall |
| --- | --- |
| m-language | long, lamba(Hi), lāmb(Ma), labi(Gu), long(Fr), lang(Sw) |
| Non-members | dugo – Baltic and Slavic languages use words cognate with deergha. fada(Irish), makrys(Greek) |
| m-alphabet(core) | l, n, m, b, g, a, o, i |
| m-alphabet (Extended) | NA |
| Remarks | Here Indian Languages have direct cognates with European Languages. Sanskrit tends to use Deergh. However Sanskrit word vilamb(delay) indicates Sanskrit origin of the above words. |

Table 10: m-language for word group “High”

| Theme | High |
|---|---|
| m-language | unc(Hi), ucca(Sa), ucca(Be) hoch(Ge), hoog(Du) hog(Sw), Haut(Fr) |
| Non-members | Uyar(Ta) |
| m-alphabet(core) | {u, c } |
| m-alphabet (Extended) | {u, n, c, t, g, a, u, e} |

Table 11: m-language for word group “Below, Low, Lowly”

| Theme | Lowly/below |
|---|---|
| m-language | Lowly:nīc(Sa), Below: nīce(Hi), nizhe(Ru) nizsie(Sl) |
| Non-members | Many |
| m-alphabet(core) | n, c |
| m-alphabet (Extended) | n, c, ī, e, zh, s |

Next, we analyze the Dravidian Language words using sounds. In Table 12 below, we analyze how the words for numbers are constructed in Dravidian Languages. There are sound shifts from pa to ha (Pattu and Hattu) in Kannada. The ‘b’, ‘p’, and ‘v’ sounds also seem to be used interchangeably. Malayalam and in some cases, Tamil manage without the suffix ‘u’, whereas others customarily use it.

Table 12: Words for numbers in Dravidian Languages

| **Number** | **Kannada** | **Tulu** | **Telugu** | **Tamil** | **Malayalam** | **m-alphabet (Extended)** | **m-alphabet (core)** |
|---|---|---|---|---|---|---|---|
| One | ondu | onji | okati | onru | onn | o,n,d,u,j,I,k,a,t,r | o,n |
| Two | eraḍu | ra*dd* | ranḍu | iran*d* | ran*d* | e,r,a,d,u,n,i | r,a,*d* |
| Three | mooru | mooji | muḍu | munr | munn | m, ū,r,u,j,I,d,r | m, ū |
| Four | nālku | nāl | nālugu | nānku | nal | n,ā,l,k,u,g,n | n,ā,l |
| Five | aidu | ain | aidu | aintu | anj | ai,d,u,n,t,a,j | ai,n |
| Six | āru | āji | aru | āru | ār | ā,r,u,j,i | ā,r |
| Seven | elu | el | edu | elu | el | e,l,u,d | e,l |

| Eight | entu | edma | enimidi | ettu | ett | e,n,t,u,d,m,ā,I,d | e,t |
|---|---|---|---|---|---|---|---|
| Nine | ombattu | ormbā | tommidi | onpatu | ompat | o,m,b,a,t,u,r,ā,d, n,p | o,m,t |
| ten | hattu | patt | padi | pattu | patt | h,a,t,u,p,d | p,a,t |
| twenty | ippattu | irva | irvai | irupat | irupat | I,p,a,t,u,r,v,i | I,r,v,p,a,t |
| thirty | muvattu | muppa | muppai | muppatu | muppat | m,u,v,a,t,p | m,,u,p,a,t |
| fourty | naluvattu | nālpa | nalabhai | narpatu | nalpat | n,ā,l,u,v,a,t,u,p,bh,r | n,ā,l,p |
| fifty | aivattu | aiva | yabhai | aimpatu | ampat | ai,v,a,t,u | ai, v, p, |
| Phonemic Affinity | u, v, d | j,ā | *d,bh* | n, r | a, n, m | | |
| Excluded Phonemes | | | | v | v | | |

## 3.3. Linguistic Analysis using Finite State Machines

Pānini's method of understanding the language consists of

- Breaking the sentence into words
- Words into Prakriti (original part) and Pratyaya (suffix).
- Further break Prakriti into components if possible and needed.
- These components are repeatedly seen in multiple words
- Map these repeating components with repeating meanings
- Assigning meanings to these components
- Also observe how these meanings in a sentence are connected

Pānini's method of analyzing words consists of

- Observing the repeated occurrences of letters or groups of letters in different words
- Observe the repetition of the same meaning in different words
- Map repeating sounds with repeating meanings.
- Assigning meaning to the component of a word.

This process results in deriving common Dhātus (root words) out of the Prakriti component and identification of Pratyayas (common suffixes) that get attached to multiple words depending on the meaning to be conveyed. Pānini ordains a step-by-step process for joining the Prakriti and Pratyaya. Phonetic and intonation changes when words come together (Sandhi and Samāsa) also need to be considered.

The methodology we propose builds on these foundational concepts.

### 3.4 Proposed Methodology

In this paper, we propose the following methodology.

- We construct a phonetic map using Pānini's System of sounds.
- We represent sounds and words including parts of words under construction as states and represent each word as a state-transition diagram.
- Construct a unified state transition diagram for words belonging to a word group with associated m-language and m-alphabet. Here a completed word is represented as an accepting state.
- Compute distances on the phonetic map, each word traverses as it gets constructed. Compute inter-word distances for word groups. This can be useful to identify central words or original words that have led to other words.
- Associate a grammar (NT, T, P, S) where NT is a set of non-terminals, T is a set of Terminal Symbols, S is the starting Symbol, and P is a set of production rules, with each m-language.
- Derive a Finite Automaton that accepts words that belong to a given m-language.
- The m-languages can be expanded to include groups based on ontological considerations when words express related concepts and grammatical considerations when words are used to convey related constructs.
- The Finite Automata can be extended to accommodate suffixes that also have commonality across languages as well as undergo transformation within languages.

Once we have a repository of m-languages we can derive additional words and discover linkages between words that were not widely known. The overall idea is to analyze words beyond the confines of individual languages and improve their intelligibility without necessarily requiring one to know the corresponding language in its entirety. The proposed approach can enable us to appreciate how the words change over temporal, geo-spatial, cultural, religious, professional locales, landscapes, and milieu.

Here we have used Google Translate (translate.google.com) extensively. We also have used dictionaries (learn.sanskrit.com) and our knowledge of languages as native speakers.

### 3.5 Proposed Phonetic Map of Sounds

First, we lay out a geometric space of sounds as per Pānini's System of Sounds. This is used to create the phonetic map. In this map, each word is a path traversed. Comparing two words is a matter of comparing two paths. Words with common roots may be naturally represented as they share the first part of the word. Words that have sound shifts may show divergence only at those points where the shift has happened. Figure 3 illustrates the proposed Phonetic Map.

The topology of the map, we have constructed using the following thought process. Origin is when no sound is produced and no effort is exercised. On the Y axis, lower coordinates are given for vowels and higher Coordinates are given for consonants. The semi-vowels are accommodated next to vowels. Sibilants and aspirates are accommodated just before consonants. On the X-axis, the velar sounds have low coordinates and labial sounds have higher coordinates. Thus, we have depicted the voice box on the left bottom extreme and the mouth at the right bottom extreme. Then among consonants, we have given a lower X coordinate for an unaspirated sound and a higher coordinate for the aspirated sound. The voiced sounds are placed higher compared to unvoiced sounds.

Nose

| 17 | ङ | | ञ | | ण(ṇ) | | न्(n) | | म्(m) | | Nasal |
|---|---|---|---|---|---|---|---|---|---|---|---|
| 16 | | घ्(gh) | | झ्(jh) | | ढ(ḍh) | | ध(dh) | | भ(bh) | Voiced-Aspirated |
| 15 | ग्(g) | | ज्(j) | | ड | ळ(ḷ) | द(d) | | ब(b) | | Voiced |
| 14 | | ख्(kh) | | छ्(ch) | | ठ(ṭh) | | त्(t) | | फ(ph) | Aspirated |
| 13 | क्(k) | | च्(c) | | ट(ṭ) | | त्(t) | | प्(p) | | Tenue |
| 12 | ह्(h) aspirate | | श्(ś) | | ष्(ṣ) | | स्(s) | | | | Sibilant |
| | Kantavya | | Talavya | | Murdhva | | Datavya | | Austa | | |
| | Guttaral | | Palatal | | Cerebral | | Dental | | Labial | | |
| 11 | व्(v) | | | | | | | | | | Semi-vowels |
| 10 | ल्(l) | | | | | | | | | | |
| 9 | र्(r) | | | | | | | | | | |
| 8 | य्(y) | | | | | | | | | | |
| 7 | अ(a) | आ(ā) | इ(i) | ई(ī) | ऋ | ॠ | ऌ | ॡ | उ(u) | ऊ(ū) | Vowels |
| 6 | ऐ(ai) | | | | | | | | | | |
| 5 | ए(e) | | | | | | | | | | |
| 4 | औ(au) | | | | | | | | | | |
| 3 | ओ(o) | | | | | | | | | | |
| 2 | ं: (am) | | | | | | | | | | |
| . 1 | ः(ah) | | | | | | | | | | |
| 0 | 1 | 2 | 3 | 4 | 5 | 6 | 7 | 8 | 9 | 10 | |

Vocal Box

Mouth

Figure 3: Phonetic Map of Indic Sounds (Devanagari)

Certain vowels are considered a combination of basic vowels. For example, we consider sound ai gets constructed due to the quick succession of sounds 'a' and 'i'. Then we consider sound e is composed due to the combination of sounds 'a' and 'i'. Similar considerations apply to au and o sounds which make use of 'a' and 'u' sounds.

Alternative topologies also may be considered where labials get low X-coordinates and velars get high X-coordinates. In such as case, the distance from the origin may be a better indicator of the effort required to generate a sound. However, the present layout, we feel is acceptable and easier to relate to. Next, we tabulate the coordinates of sounds on the phonetic map in tables 13-15. Table 16 contains examples of words.

Table 13: Vowel Sounds

| Sound | Coordinate | Sound | Coordinate | Sound | Coordinate |
|---|---|---|---|---|---|
| अ | *(7,1)* | आ | (7,2) | इ | (7,3) |
| ई | (7,4) | ऋ | (7,5) | ॠ | (7,6) |
| ऌ | (7,7) | ॡ | (7,8) | उ | (7,9) |
| ऊ | (7,10) | ऐ | (6,2) | ए | (5,2) |
| औ | (4,5) | ओ | (3,5) | ं | (2,5) |
| ः | (1,1) | | | | |

Table 14: Consonant Sounds

| Sound | Coordinate | Sound | Coordinate | Sound | Coordinate |
|---|---|---|---|---|---|
| क् | (13, 1) | ख् | (14, 2) | ग् | (15,1)) |
| घ् | (16,2) | ङ | (17, 1.5) | च् | (13,3) |
| छ् | (14,4) | ज् | (15,3) | झ् | (16,4) |
| ञ | (17, 3.5) | ट | (13,5) | ठ | (14,6) |
| ड | (15,5) | ळ | (15,6) | ढ | (16,6) |
| ण | (17,5.5) | त् | (13,7) | थ | (14,8) |
| द | (15,7) | ध | (16,8) | न् | (17, 7.5) |
| प् | (13,9) | फ | (14,10) | ब | (15, 9) |
| भ | (16, 10) | म् | (17, 9.5) | | |

Table 15: Sibilants and Semivowels

| Sound | Coordinate | Sound | Coordinate | Sound | Coordinate |
|---|---|---|---|---|---|
| श् | (12, 3.5) | ष् | (12,5.5) | स् | (12, 7.5) |
| ह् | (12, 1.5) | य् | (8, 2.5) | र् | ((9, 3.5) |
| ल् | (10,4.5) | व् | (11, 5.5) | | |

Table 16: Word Examples

| Word | Path | Word | Path |
|---|---|---|---|
| kapi | (13,1) (7,1) (13,9) (7,4) | hrudaya | (12,1.5) (9, 3.5) (7,1) (15,7) (7,1) (8,2.5), (7,1) |
| ape /eip/ | (5,2) (13,9) | heart /ha:t/ | (12,1.5) (7.2) (9,3.5) (14,8) |
| go | (15,1) (3,5) | mana | (17,9.5) (7,1) (17,7.5) (7,1) |
| cow/kau/ | (13,1) (4,5) | mind mʌɪnd/ | (17,9,5) (6,2) (17,7.5)(15,7) (7,1) |
| bo | (15,9) (3,5) | mental /ˈmɛnt(ə)l | (17, 9.5) (5,2) (17,7.5) (13,7) (7,1) (10,4.5) |

In the above table, it can be argued that the English word mental is closer to the Sanskrit word mana rather than 'mind'. In the case of hrudaya, 'hrut' is the root word that is close to the heart as well. The Irish word 'bo' is the word for cow. This may be unrelated, but it ends with the same vowel sound as go, the Sanskrit word for cow. The old English word for cow is coo. English uses the word bovine as a generic term to mean "affecting cattle". The German word for cow is 'kuh'. Persian has retained go. Latvian also has retained 'govs'. Otherwise, most European Languages use words starting from k for the cow. In contrast, when it comes to interrogatives, Sanskrit and Indian Languages as well as most European Languages, use words starting with the 'k' sound whereas Germanic languages use words such as who and 'hvem'. The etymological analysis of the word wheel also leads one to a root starting with 'k'. Thus, which word is original can become a matter of debate and controversy.

The sounds which are not included in Pānini's System of Sounds such as Alveolar or fricative sounds can be given intermediate coordinates on the phonetic map.

### 3.6 Finite State Machine Preliminaries

A state machine consists of states and transitions. There may be one or more initial states and one or more terminal states. From the terminal States, no further transitions happen. There can be transitions back to the same state as well. Figure 3 below illustrates a state machine. Here S1, S2, S3, and S4 are states represented by circles, and T1, T2, T3, and T4 are transitions depicted using arrows. S1 is the start state. S4 the terminal state is represented using a donut-shaped circle. The transitions happen from state to state depending on the input given to the system in a particular state. Figure 4 below depicts a finite state machine.

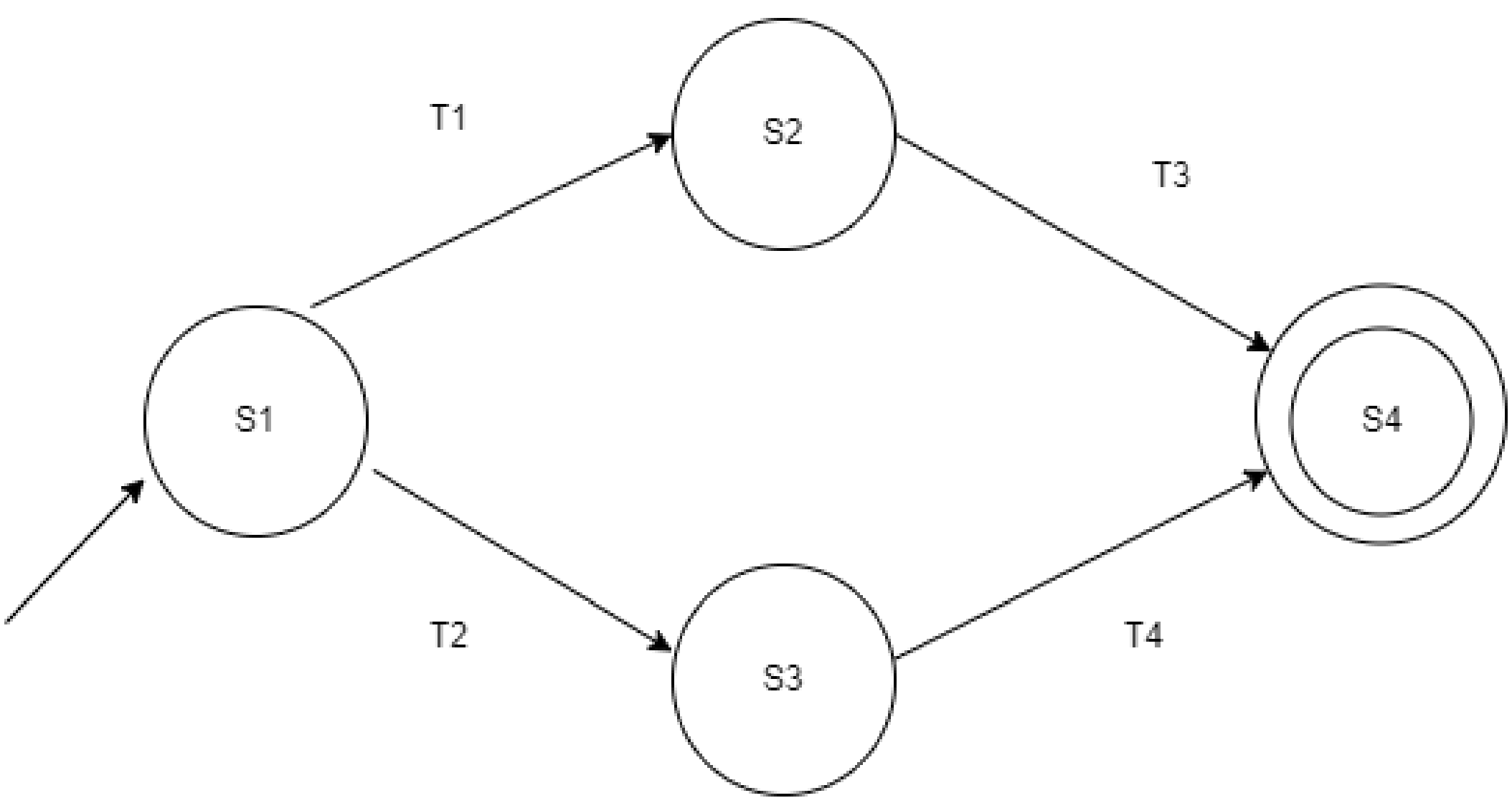

Figure 4: Finite State Machine

A Finite Automata is a State Machine that takes a string of symbols as input and changes the state accordingly. For a given input, the automaton can move to another state or remain in the same state. After processing a symbol string if the Automaton reaches an accepting state, then it has accepted that string as a valid string. One can also configure bad states, where from a given state when a particular input symbol is encountered it will reach the bad state when the string is rejected. There are two kinds of Finite Automata: Deterministic and Non-deterministic. Here a string w=a1a2…an, where a1, a2, … are input symbols.

A Deterministic Finite Automata (M) is a Quintuple

$M= (Q, \sum, \delta, qo, F)$

Q: a finite set of states

q0: Start State, where q0 € Q.

$\sum$: a finite set of input symbols

F: final states where $F \subseteq Q$

$\delta$: Transition function where $\delta: Q \times \sum \rightarrow Q$

The language accepted by DFA M is

$L(M) = \{w \mid \delta\text{^} (q0, w) € F\}$

If for a given input, more than one kind of transition happens such an automaton is non-deterministic. If for a given input, there is no clarity on what happens such automata are non-deterministic. Finite automata with multiple start states are non-deterministic. Thus, only automata with a single start state and a uniquely defined transition for every input are considered Deterministic.

The most basic and foundational construct for processing symbols is the Atomic Proposition. Here AP is a set of Atomic Propositions and AP-INF is a set of infinite words over Power Set (AP). A set of words is termed as language. To form words, one needs an alphabet. For example, let us say (a, b) is the alphabet. Then, a formal/rule-based language can accept only a's, only b's or a's and b's alternating. In the case of a language that takes only a's as input, when we model it as a finite automaton, the initial and end-states are the same. In this case, since there is no transition defined when the input is b, it is considered a Non-deterministic Finite Automaton. Figure 5 below shows an automaton that accepts only 'a' as the input. Here 'a', 'aa', and 'aaa' are the words of the language.

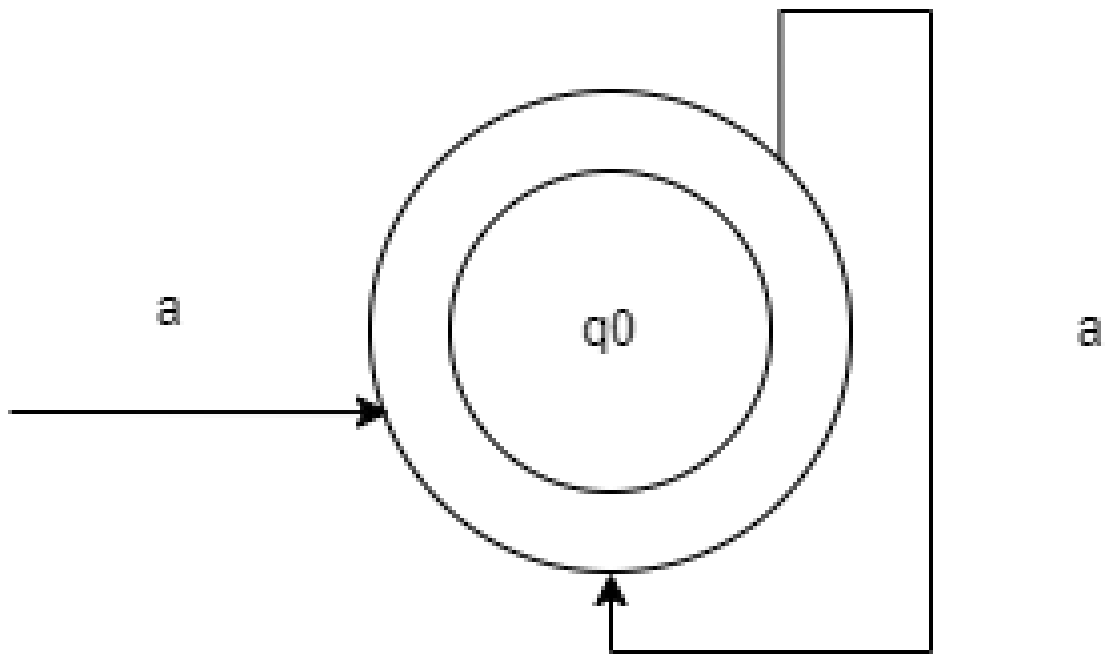

Figure 5: Finite Automaton which accepts only "a"

Thus, we have:

Alphabet {a,b}

Language a* = { ѐ, a, aa, aaa, aaaa, $a^5$ , …}

Another example of Language using the same alphabet is

L1 = {ѐ, ab, abab, ababab, … }

Here ѐ is an empty symbol and a word of length 0. The language accepts alternating 'a's and 'b's or empty symbols.

The following finite automaton illustrates a language where the initial symbol is a, and one or more b's. Figure 6 illustrates the same. The language

L2= {a, ab, $ab^2$, $ab^3$,… }

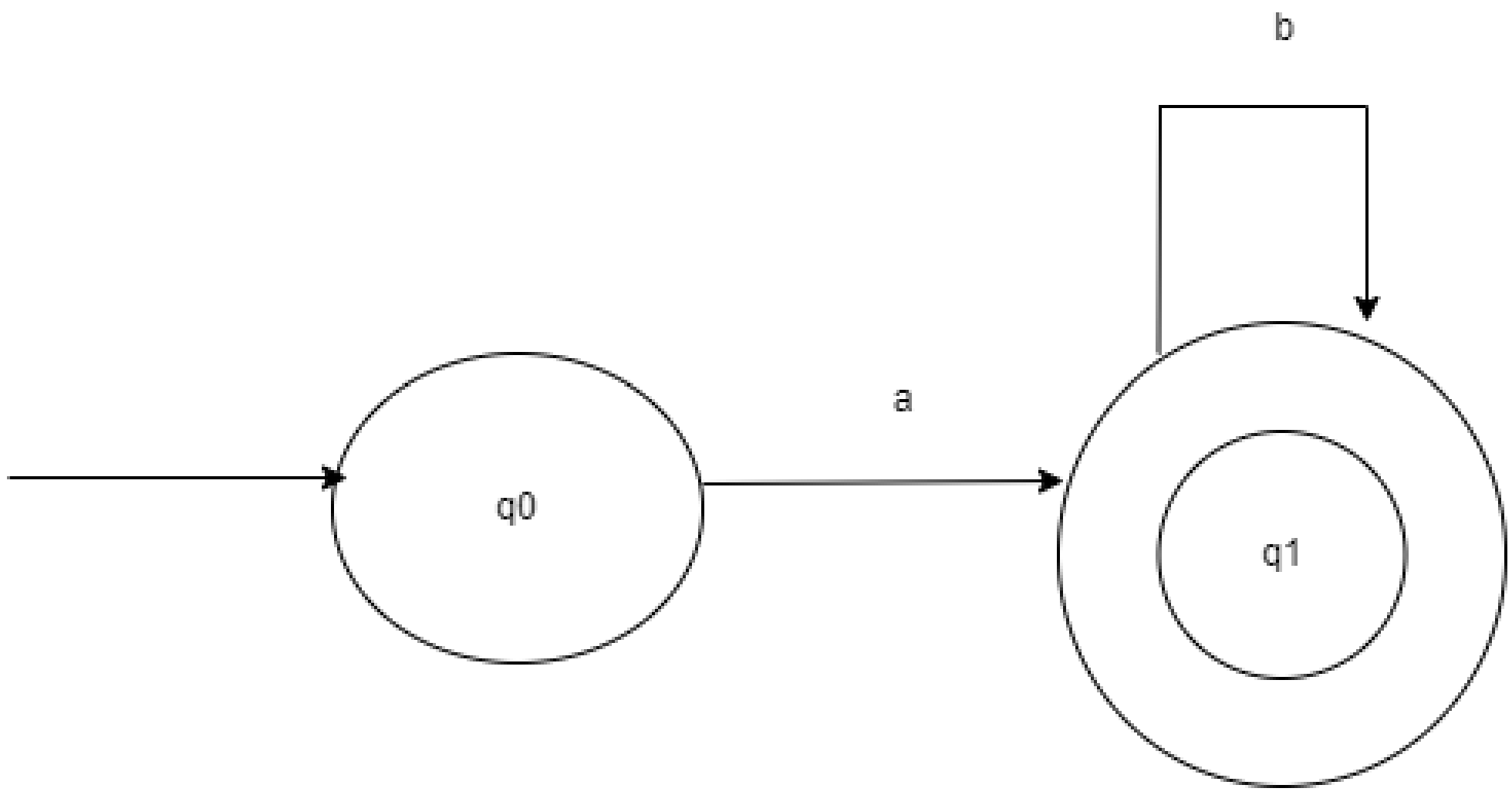

Figure 6: Finite Automaton that accepts a and then one or more b's

For example, if ∑.is alphabet, ∑* is the set of all words over ∑, a word starting with 'a' and ending with 'a' can be represented as a∑*a.

The languages that are accepted by finite automata are called regular languages and for every regular language, there is a DFA that accepts it. Every NFA (Non-deterministic Finite Automaton) can be converted to an equivalent DFA (Deterministic Finite Automaton).

### 3.7 Application of Proposed Methodology

We take a group of words that relate to each other phonetically, semantically, grammatically, and/or ontologically. This we call m-language and give it a unique identifier. The sounds that are used in constructing the words of the m-language constitute m-alphabet. This analysis and construction of m-language requires reasonable knowledge about the words and languages involved. At the same time, the process of analysis itself can be educative. We can extend the m-language and cover related concepts. In certain languages, by adding specific sounds at the beginning of a word, we end up with an antonym.

Next, we look at representative cases. In the following m-language, we address the poetry theme. Here starting phoneme is common. Figure 7 illustrates the state transition diagram where each phoneme as well as word under construction are states. The completed word is accepting state.

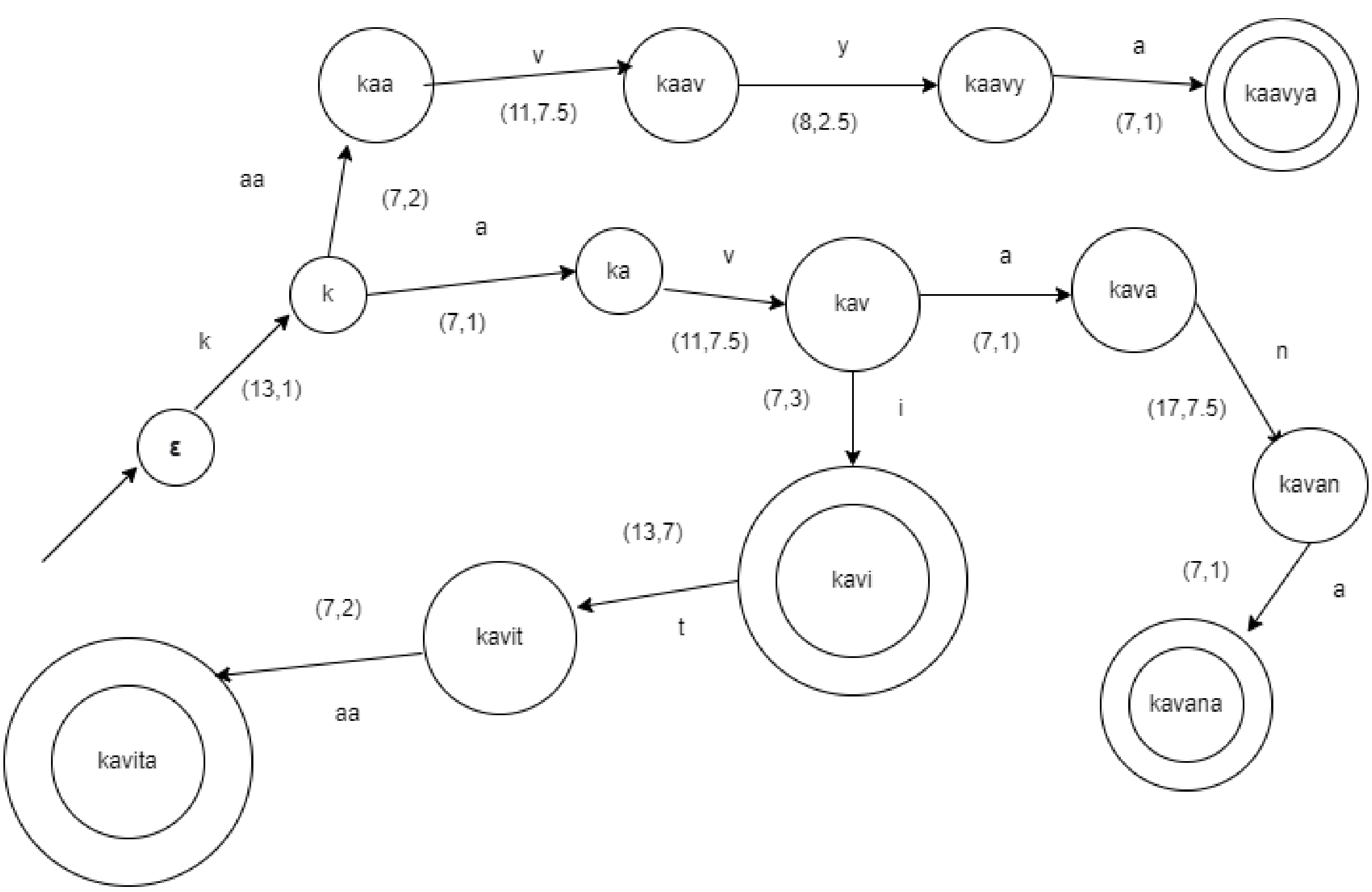

Figure 7: State Transition Diagram for words related to Poetry Theme

Here we have represented Kavi(poet), Kavitā(poem), Kavana(poem), Kāvya(Epic in poetic form), and Kavana(poem). The last word is found only in Kannada. Other words are common across Indic languages. With each m-alphabet, we associate the coordinates on the phonetic map covered in the last section. Thus, corresponding

m-language = { kavi, kavitā, kāvya, kavana}

m-alphabet = { k,v,t,y,n,a,ā,i} = {(13,1), (11,5.5), (13,7), (8,2.5), (17,7.5), (7,1), (7,2) (7,3)}

Here k and v are basic alphabets that are extended to make new words. Here basic sounds remain the same and new word forms are due to grammar. The way sounds were associated with coordinates on the phonetic map, the combination of sounds and words can be associated with phonetic distances that traverse. Table 17 illustrates the method used to compute distances for states. We express distance as X and Y components.

Table 17: Words with Poetry theme

| Input and Coordinates | | | State and Manhattan Distance | | | Input and Coordinates | | | State and Manhattan Distance | | |
|---|---|---|---|---|---|---|---|---|---|---|---|
| Null | 0 | 0 | Null | 0 | 0 | Null | 0 | 0 | Null | 0 | 0 |
| k | 13 | 1 | k | 13 | 1 | k | 13 | 2 | k | 13 | 1 |
| a | 7 | 1 | ka | 19 | 1 | a | 7 | 1 | ka | 19 | 1 |
| v | 11 | 5.5 | kav | 23 | 5.5 | v | 11 | 5.5 | kav | 23 | 5.5 |
| i | 7 | 3 | **kavi** | **27** | **8** | a | 7 | 1 | kava | 27 | 10 |
| t | 13 | 6 | kavit | 33 | 12 | n | 17 | 7.5 | kavan | 37 | 16.5 |
| ā | 7 | 2 | **kavita** | **39** | **17** | a | 7 | 1 | **kavana** | **47** | **23** |
| Null | 0 | 0 | Null | 0 | 0 | | | | | | |
| k | 13 | 1 | k | 13 | 1 | | | | | | |
| ā | 7 | 2 | kā | | 2 | | | | | | |
| v | 11 | 5.5 | kāv | 23 | 5.5 | | | | | | |
| y | 8 | 2.5 | kāvy | 26 | 8.5 | | | | | | |
| a | 7 | 1 | **kāvya** | **27** | **10** | | | | | | |

Next, we can tabulate inter-word distances. See Table 18 below.

Table 18: Inter-word distances Poetry Theme

| | Kavi | Kavita | Kāvya | Kavana | Row Sum |
|---|---|---|---|---|---|
| Kavi | 0,0 | 12,9 | 0,2 | 20,15 | 32,26 |
| Kavita | 12,9 | 0,0 | 12,7 | 8, 6 | 32, 15 |
| Kāvya | 0,2 | 12,7 | 0,0 | 20,13 | 32, 22 |
| Kavana | 20,15 | 8,6 | 20,13 | 0,0 | 48,34 |

The above analysis alludes to the possibility that Kavita and Kāvya are central words. Kavi here is the most basic word. We can repeat the same analysis by excluding Kavana. Here Kāvya is more central than Kavita.

Table 19: Inter-word distances Poetry Theme excluding Kavana

| | Kavi | Kavita | Kāvya | Row Sum |
|---|---|---|---|---|
| Kavi | 0,0 | 12,9 | 0,2 | 12, 11 |
| Kavita | 12,9 | 0,0 | 12,7 | 24, 16 |
| Kāvya | 0,2 | 12,7 | 0,0 | 12,9 |

For the above case, Figure 8 below illustrates the Deterministic Finite Automata, which we term Morphological Finite Automata (MFA). Here Q0 is the starting symbol, Q5, Q7, Q11, and Q4 are accepting states. We have made use of null symbols to end with an accepting state and continue to form more words in parallel. Along with the word, in the parenthesis, the language is indicated.

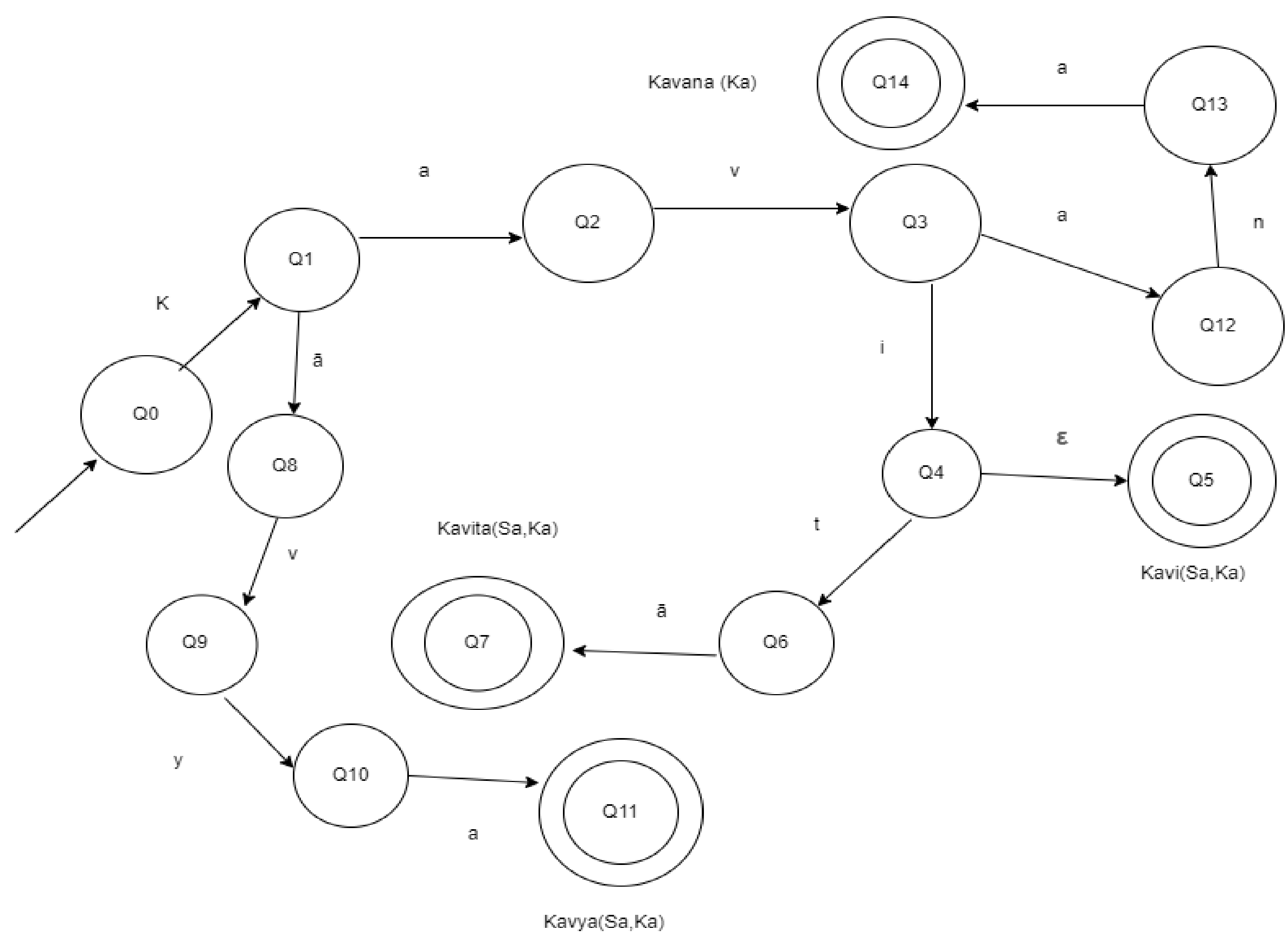

Figure 8: MFA for Kavita and related words

Corresponding to the above MFA, the production rules for the grammar can be written as follows.

Q0 ->kQ1; Q1->aQ2; Q2->vQ3;Q3->i|iQ4; Q4-> tQ6; Q6 ->ā

Q0->kQ1;Q1->āQ8; Q8->vQ9;Q9->yQ10; Q10->a

Here tā and ya are standard and commonly used suffixes in Indian Languages. The production rules can be rewritten as follows by accommodating the suffixes as terminal symbols in their own right. Similar words are Savita, Kartavya, etc.

Q0 ->kQ1; Q1->aQ2; Q2->vQ3; Q3->iQ4->tā

Q0->kQ1; Q1

->āQ8; Q8->vQ9; Q9->ya

m-language(L) = {S->* W, W is related to Poetry Theme}

Below we look at words that mean “the well’, cutting across languages. Sanskrit uses Koopa for a deep well and Vapi for a broad well. Figure 9 below depicts the corresponding MFA.

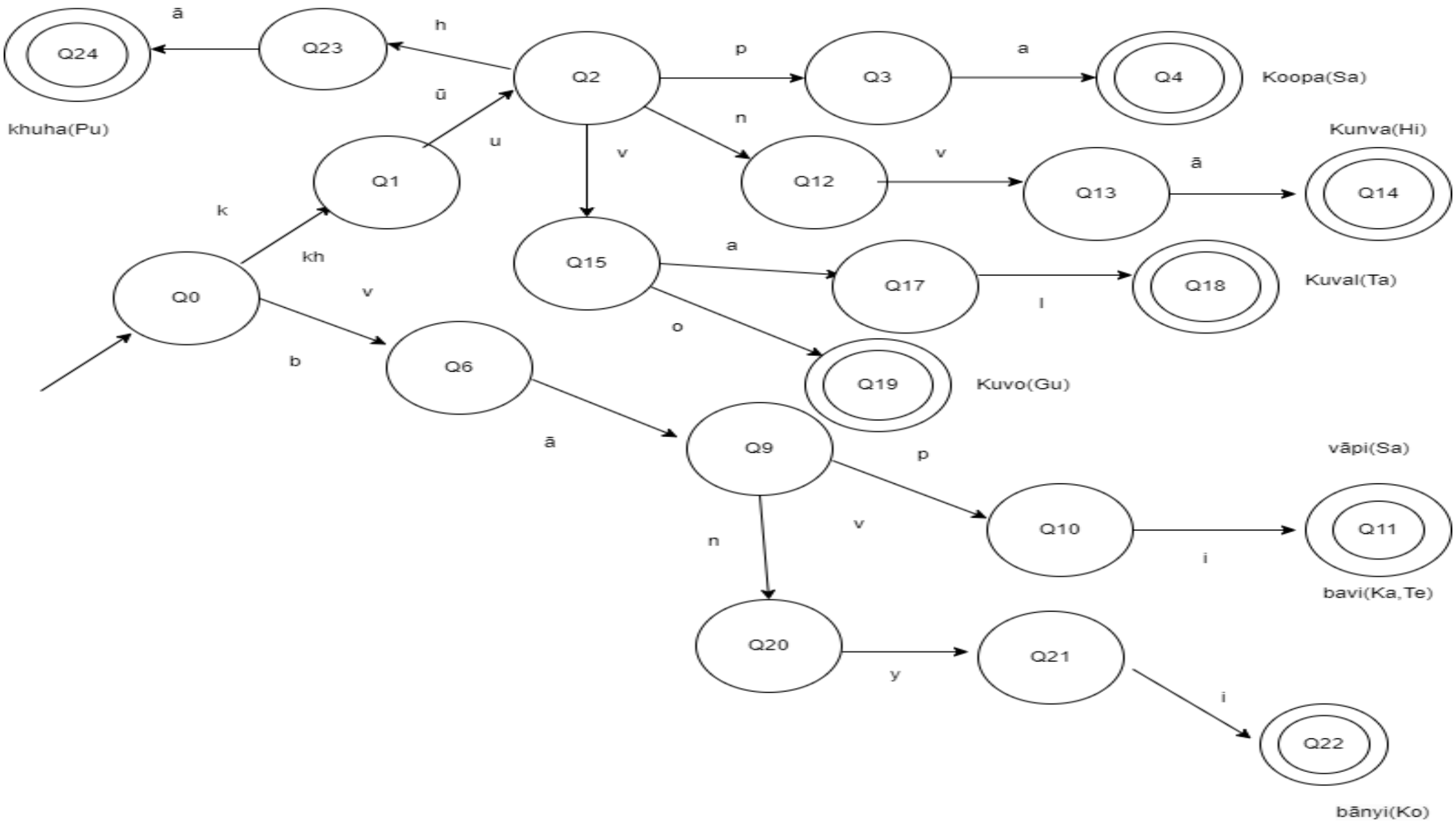

Figure 9: MFA for words meaning “the well”.

The production rules can be arrived at similarly as in the previous case. Here the m-alphabet corresponding to Koopa is {k,p,v} and vowels. By adding b to the same alphabet, we can accommodate a second set of words i.e. Vāpi and Bāvi.

Next, we look at an example that also starts with a common phoneme but cuts across languages. We take up the word for God in Indo-European Languages, which starts with the sound ‘d’ in many of the languages except Germanic and Russian which uses the Bhag derivative. See Figure 10.

Corresponding m-language = {deva, devs, dio, dia, theos, dieu, devaru, devudu}

m-alphabet = {d, th, a, i, u,o, s, d, r}

Greek is using “th’ sound with coordinate (14,8) instead of ‘d’ sound with coordinate (15,7). Both sounds are dental. Other than that, sounds used are nearly the same. The ‘s’ sound is used for plurals in Vedic Sanskrit and Indo-European Language. In Kannada and Telugu, the word for God is in the plural form and they use the ‘r’ and retroflex ‘*D*’ sounds respectively

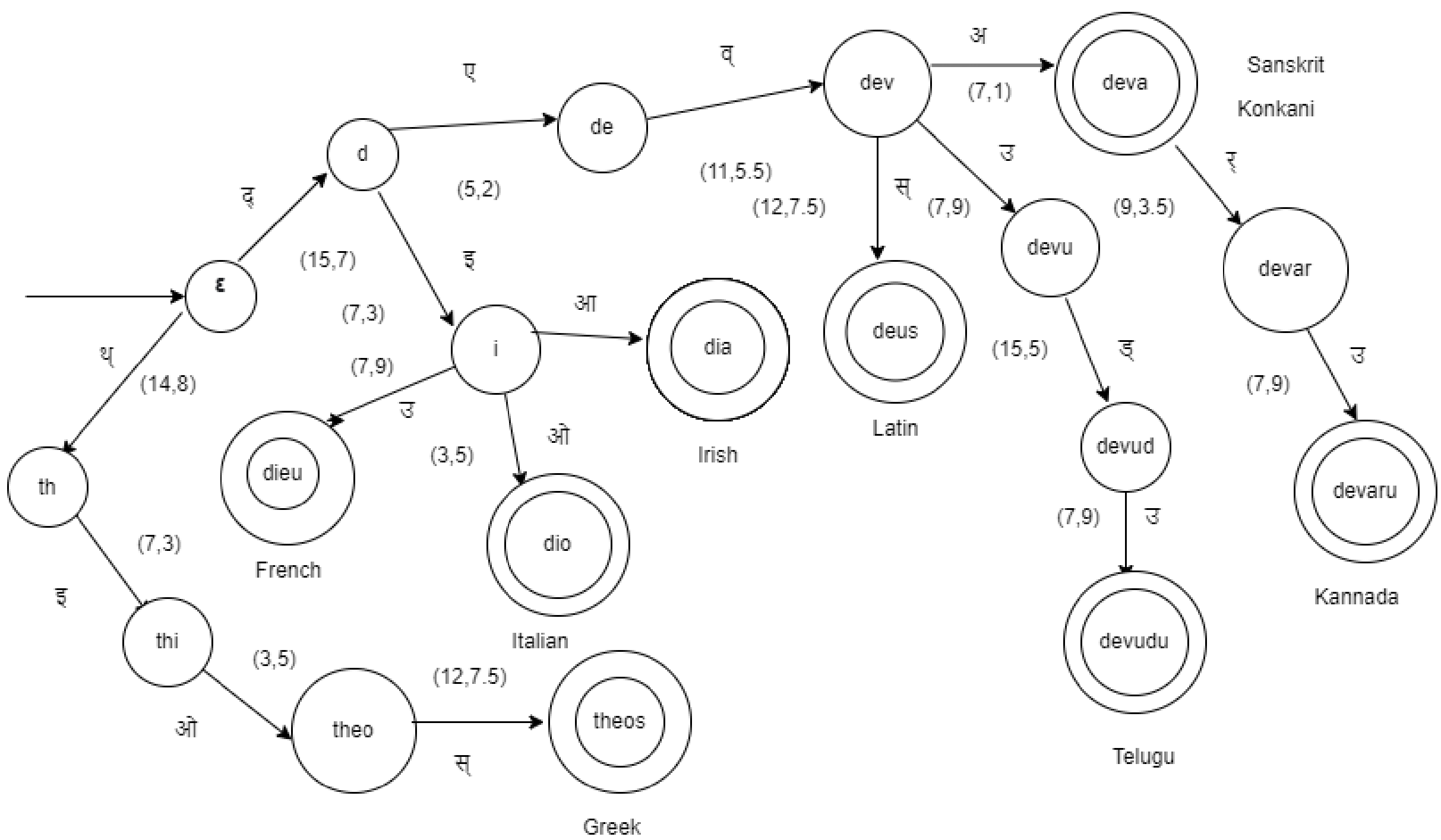

Figure 10 State Transition Diagram for words cognate with Deva

The state computation diagram for the MFA in Figure 7 is given in Table 20 below.

Table 20: Distances on Phonetic Map for Words with Sanskrit Deva

| deva | deu | dio | dia | devs | theos | divine(davain) |
|---|---|---|---|---|---|---|
| 35, 4.5 | 23,6 | 27,2 | 23,2 | 35,3.5 | 29,4 | 43, 12/5 |

The corresponding inter-word distances are given in Table 21 below.

Table 21: Inter-word Distances words cognate with Deva

| | deva | deu | dio | dia | devs | theos | Row Sum |
|---|---|---|---|---|---|---|---|
| deva | 0,0 | 12,1.5 | 8,2.5 | 12,2.5 | 0,1 | 6,0.5 | 38,8 |
| deu | 12,1.5 | 0,0 | 4,4 | 0,4 | 12,2.5 | 6,2 | 34,14 |
| dio | 8,2.5 | 4,4 | 0,0 | 4,0 | 8,1.5 | 2,2 | 26,10 |
| dia | 12,2.5 | 0,4 | 4,0 | 0,0 | 12,1.5 | 6,2 | 34,10 |
| devs | 0,1 | 6,2.5 | 8,1.5 | 12,1.5 | 0,0 | 6,0.5 | 32,7 |
| theos | 6,0.5 | 6,2 | 2.2 | 6,2 | 6,0.5 | 0,0 | 26,7 |

Here 'theos' seems to be the basic form whereas 'deva' and 'deu' seem to be more refined forms. However, if you compare the distance between 'divine' and words for God, the following picture emerges. Phonetically the word ‘divine’ is rendered as ‘davain’. Table 22 below gives the distance of ‘divine’ between different words for God.

Table 22 Distance between divine and cognate words for God

| | deva | deu | dio | dia | devs | theos |
|---|---|---|---|---|---|---|
| divine | 8,8 | 20, 6.5 | 16,10.5 | 20,10.5 | 8,9 | 14, 4.5 |

The MFA for the above set of words is depicted in a compact manner below.

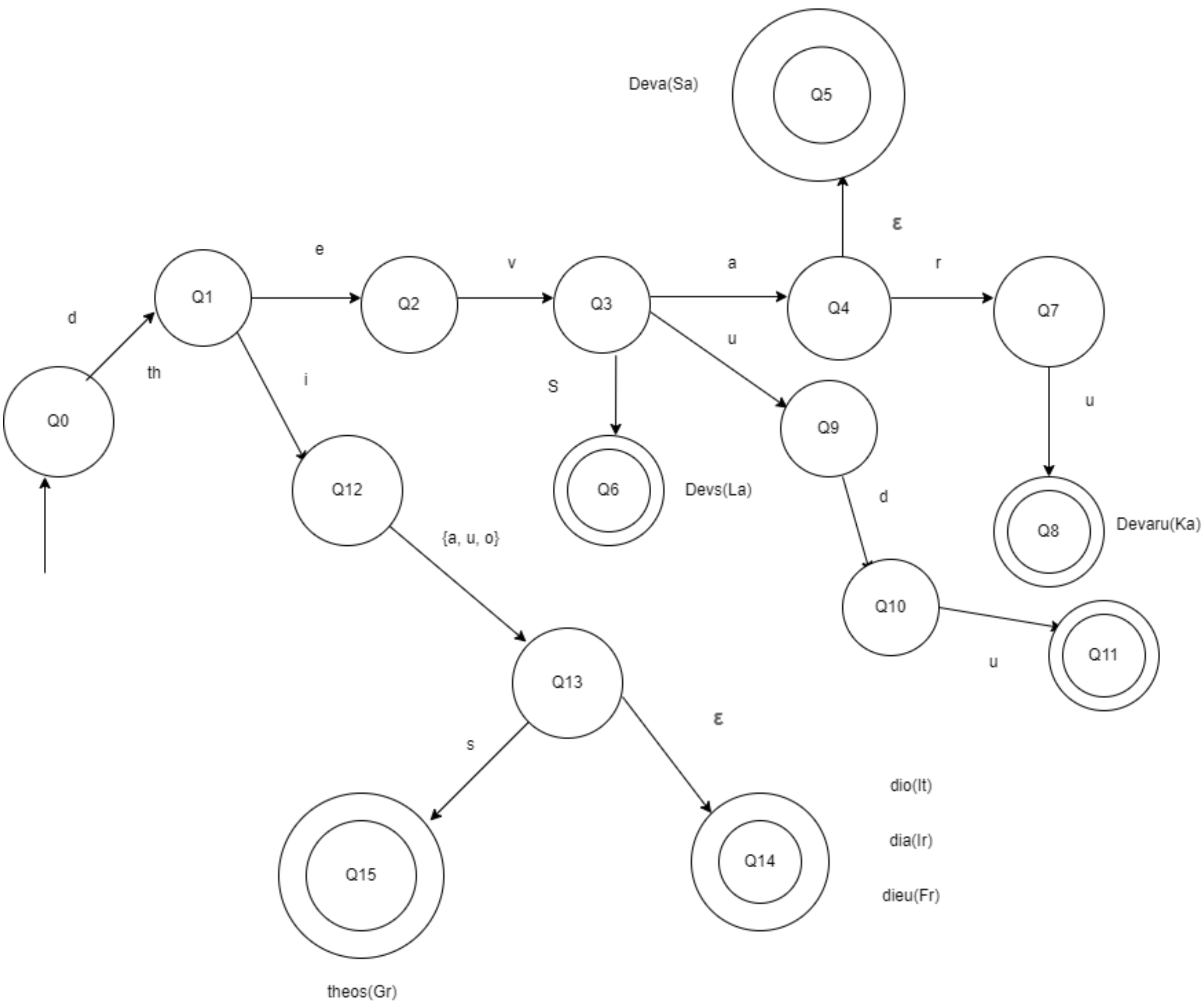

Figure 11 MFA for words cognate with Deva

The production rules in the corresponding grammar are as follows:

Q0->dQ1|thQ1; Q1->eQ2; Q2->vQ3; Q3->aQ4; Q4->Q5|rQ7; Q7->uQ8

Q1->iQ12; Q12->{a,u,o}Q13->Q14.

Q0->thQ1; Q1->iQ12; Q12->oQ13; Q13->sQ15.

Overall, our claim is that Vedic Sanskrit in prosodic form has retained the most accurate form of a word with a high degree of fidelity, while Indian and European Languages have tended to retain simpler and at times mispronounced forms in colloquial and then written forms. When you analyse a group of words (cognates and related words), the root word across languages is likely to be from Sanskrit. In India, Chandas (prosodic form) used by scholars and Bhasha (colloquial forms) used by commoners have been concurrent traditions.

Next, we look at kinship words that end with "ta" sound. These include Pitā, Mātā, Bhrātā, Duhitā, Tātā in Sanskrit. In Figure 12, we cover these and cognate words in other languages and illustrate the State Transition Diagram.

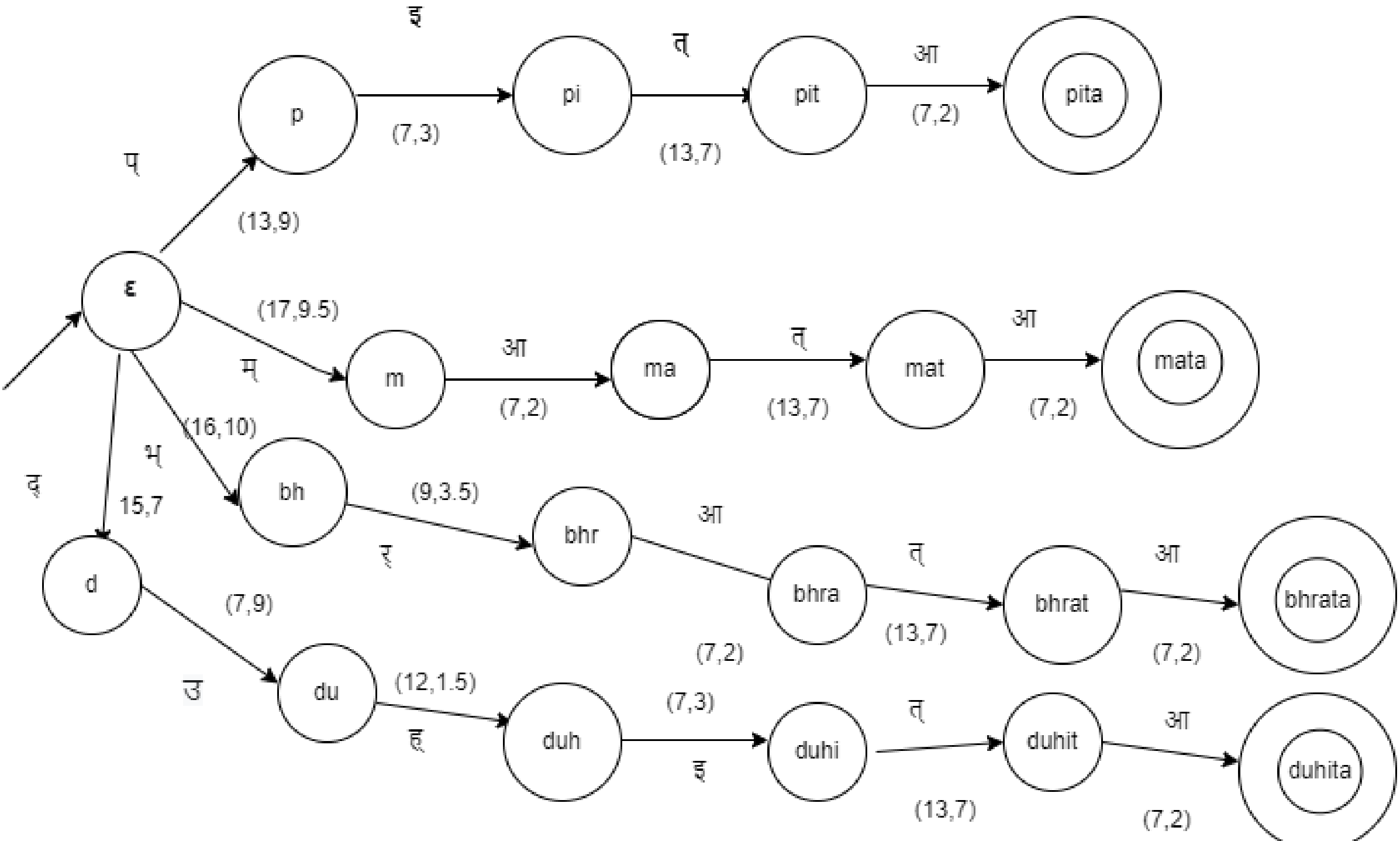

Figure 12: Kinship words ending with "ta"

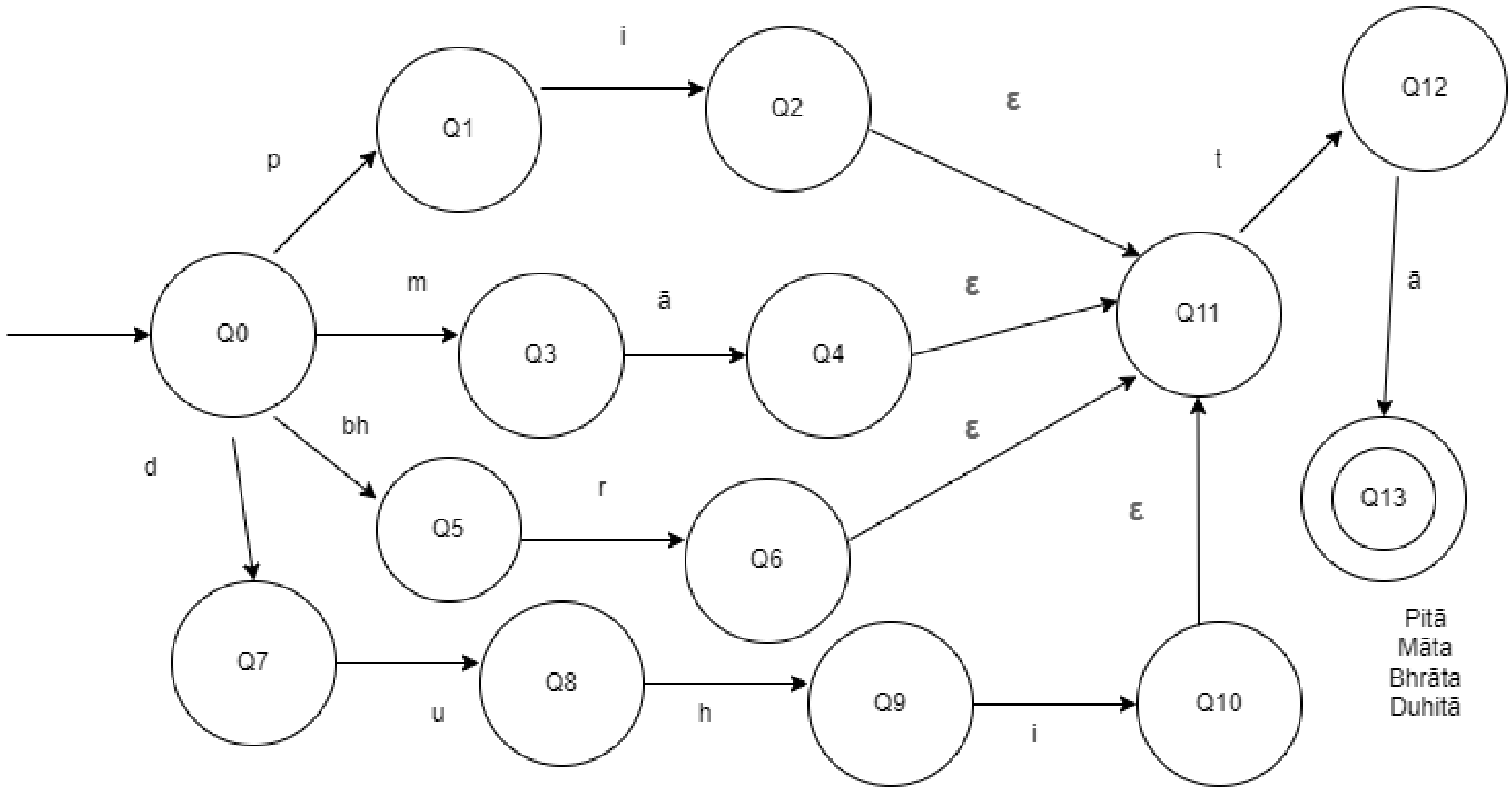

Figure 13 MFA for Kinship words ending with Ta

The corresponding MFA is illustrated in Figure 13. Here we have represented common endings by using null transitions in between.

Corresponding to the above kinship words m-language={pitā, mātā,bhrātā, duhitā} and m-alphabet = {p,m,bh,r,d,t,h,a,i,u} The state computation table for the MFA in Figure 8 is given in Table 23.

Table 23: Kinship words

| Null | 0 | 0 | Null | 0 | 0 | Null | 0 | 0 | Null | 0 | 0 |
|---|---|---|---|---|---|---|---|---|---|---|---|
| p | 13 | 9 | p | 13 | 9 | d | 15 | 7 | d | 15 | 7 |
| i | 7 | 3 | pi | 19 | 15 | u | 7 | 9 | du | 23 | 9 |
| t | 13 | 7 | pit | 25 | 19 | h | 12 | 1.5 | duh | 28 | 16.5 |
| ā | 7 | 2 | **pitā** | **31** | **24** | i | 7 | 3 | duhi | 33 | 18 |
| m | 17 | 9.5 | m | 17 | 9.5 | t | 17 | 9.5 | duhit | 43 | 24.5 |
| ā | 7 | 2 | ā | 27 | 17 | ā | 7 | 2 | **duhitā** | **53** | **32** |
| t | 13 | 7 | māt | 33 | 22 | t | 17 | 9.5 | t | 17 | 9.5 |
| ā | 7 | 2 | **mātā** | **39** | **27** | ā | 7 | 2 | ta | 27 | 17 |
| bh | 16 | 10 | bh | 16 | 10 | t | 17 | 9.5 | tāt | 37 | 24.5 |
| r | 9 | 3.5 | bhr | 23 | 16.5 | ā | 7 | 2 | **tātā** | **47** | **32** |
| ā | 7 | 2 | bhrā | 25 | 18 | | | | | | |
| t | 17 | 9.5 | bhrāt | 35 | 25.5 | | | | | | |
| ā | 7 | 2 | **bhrātā** | **45** | **33** | | | | | | |

Using the same alphabet, we can derive Pitr, Matr, Bhratr, and Duhitar which correspond to father, mother, brother, and daughter as well as Pateras. Mitera in Greek and by adding 'k' sound, Dukra in Lithuanian. Other cognate words for daughter are Dushterya(Bulgarian), Doch (Russian), and Dcera (Slovak). Among Indian languages, only Duva (Konkani), Dhi (Punjabi), Dikari (Gujarati), and Diyania(Sinhala) have retained the word. In Gujarati, Dikara(son) is related to the word for daughter Dikari. Incidentally, Dikari(Gujarati) and Dukra (Lithuanian) sound similar. Nepali uses Chori (word for a girl used for daughter) sounds akin to Corka(Polish). Many Indian Languages use Chokri. Here Romance Languages do not seem to take part in the cognate word group related to daughter. It is commonly believed that people of Sri Lanka, originally went from Orissa. However, Sinhala language has some archaic words that are common with Konkani and Vedic Sanskrit.

The word for sister is Bhagini in Sanskrit which goes with Bhrāta and thus Indian Languages use words such as Behen (Hindi), Bahiṇi(Konkani), and Bona(Bengali). Then Sanskrit uses Svasa for sister with cognates Seusa (Lithuanian), Soror (French), and Sistra (Russian). Even Finnish has Sisko. The only exceptions are Celtic Languages and Greek which seem to use very different words. Also, unlike the common understanding that retroflexes are probably loans from Dravidian Languages, they are well-established in Konkani, Punjabi, and Marathi.

Next, we look at words for son and daughter-in-law across languages. See Figure 14.

Figure 14: Words for son and daughter-in-law

Here Sanskrit word 'sunu' has a cognate word in Germanic as well as Baltic languages but not so much in Romance languages. The concept of Daughter-in-law when interpreted as a son's wife is 'snusha' in Sanskrit. Similar constructs are Snuka (Bulgarian) and Soon/Suna(Konkani) Words Nuha(Punjabi), Nos(Kashmiri), Nuos(Ancient Greek) and Nora(Portuguese) seem to have commonality with the same word group  Incidentally the word in Kannada for Daughter-in-law is Sose. The state computation table for the above MFA is given in Table 24 below. Only a subset of words is represented.

Table 24: Words for son and daughter-in-law and distances

| san | sunu | sunus | son | nora | soon | snusha | snuka | nuha | sose |
|---|---|---|---|---|---|---|---|---|---|
| 27,20.5 | 37,12 | 42,12.5 | 35,12 | 39,12 | 27,12.5 | 37,16 | 39.18 | 37,17 | 37,18 |

The MFA for words meaning the daughter-in-law is shown in Figure 15.

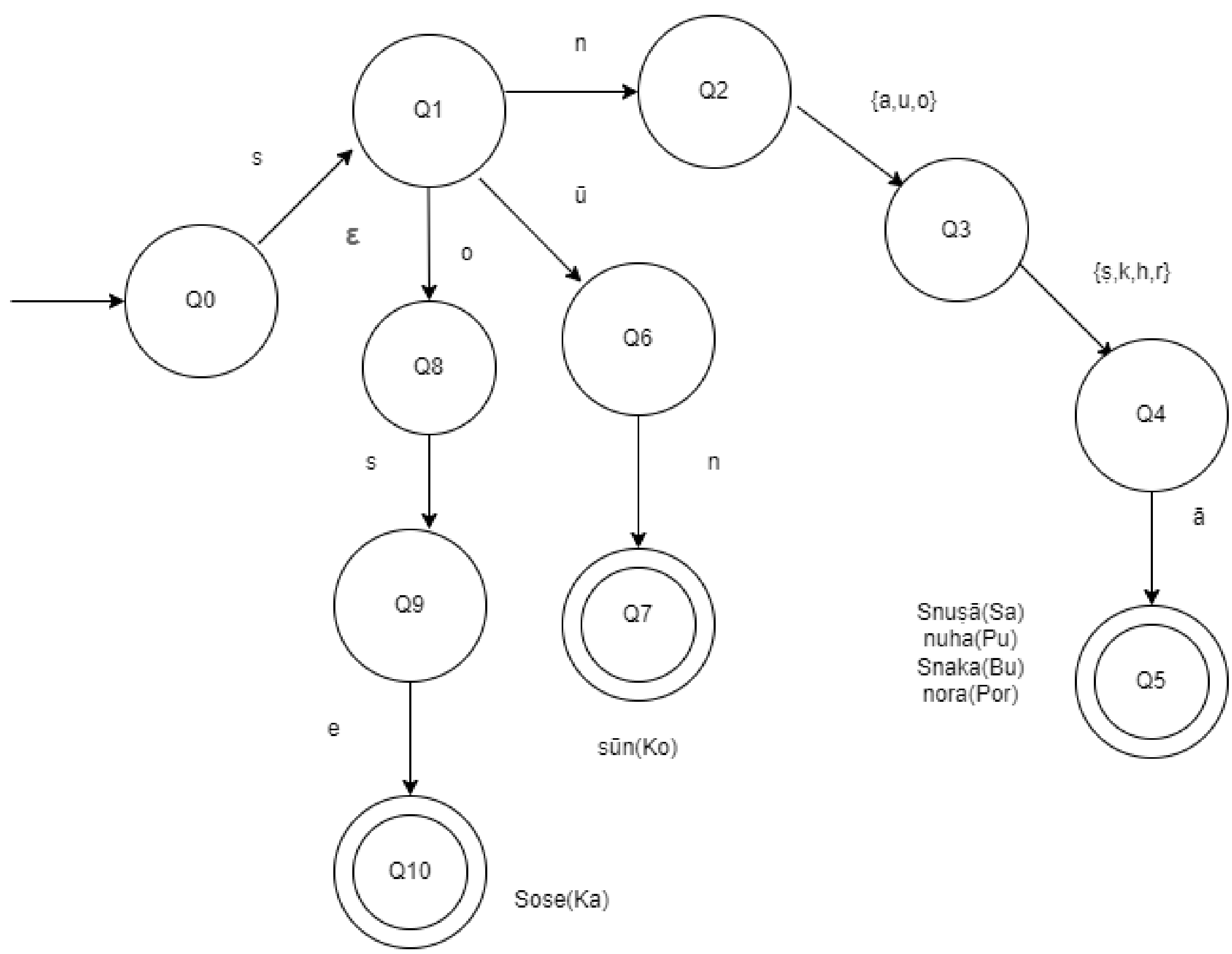

Figure 15 MFA for words meaning Daughter-in-law

Corresponding to the above MFA, basic m-alphabet ={s,n,u,a,o} Here we can consider derivations such as Snusha and Snuka as language-specific. Thus, a minor extension of m-alphabet as m-alphabet = {s, *sh, h, u,* a, k, o*,* r} can enable the generation of all the above words.

In summary, Sanskrit words in the kinship category have cognates cutting across the Indo-European Languages. The kinship word group in Sanskrit is coherent and self-contained.

Next, we look at the Apabhramsa phenomenon using the word for long. It is in Sanskrit and the corresponding word is Dīg in Konkani. Other Indian Languages either use Dīrgh as is or use some other word. Cognates are available also in Croatian, Czech, Bosnian, Macedonian, Bulgarian, Polish, Serbian, Slovak and Russian. The m-language = {Dīrgha, Deeg, Dugo, Dluho, Dulgi, Duohi, Dlugi, Dlinyy}. Here two words have same sounds but with a swap of neighbouring sounds. Thus, languages either drop r or replace r with l and arrive at the Apabramsha form. Thus, the core m-alphabet for this word = {d, g}. Sinhala old and isolated Indo-European Language has retained Digu. The words and distances on the phonetic map are given in Table 25 and the corresponding MFA is depicted in Figure 16.

Table 25: Words cognate with Dīrgha and Distances

| dīrgha | dīg | dugo | dulgi | dlugi | digu |
|---|---|---|---|---|---|
| 41,13 | 31,13 | 43,21 | 39,19 | 39,24 | 39,21 |

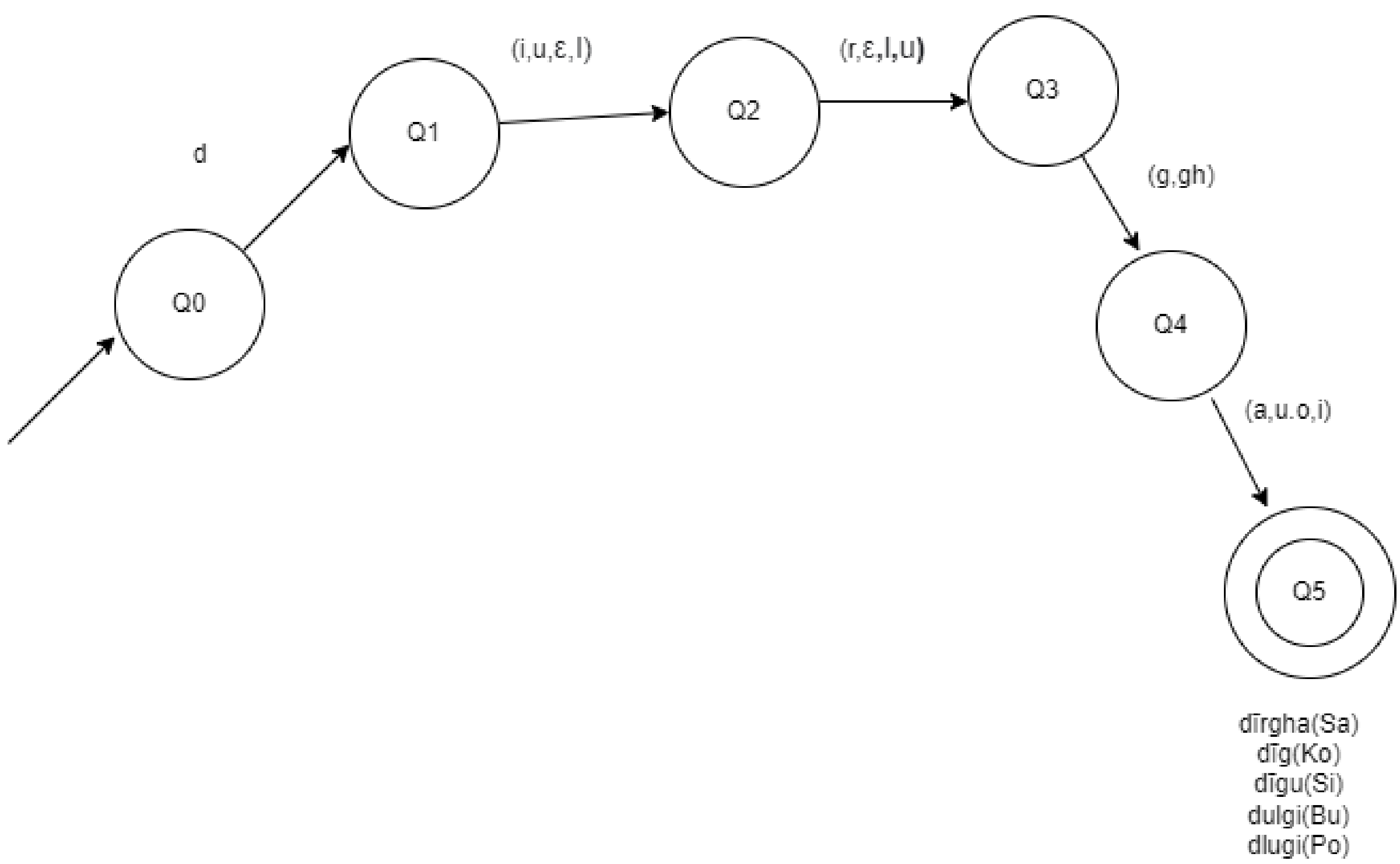

Figure 16: MFA for words cognate with dīrgha("long")

Most Indian languages use the words lamba or lambi which is closer to long in English. Both Germanic and Romance languages also use similar forms. Konkani uses lāmb to mean hang from a height (or become longer). Sanskrit uses lamb as a verb to hang/linger, with viḷamba used for delay, but the direct word for long continues to be Dīrgha. We can make a point that inter-relationships between individual Indian Languages and European Languages should also be studied. We came across a Wiktionary that attempts to derive long from 'dlogos'.

The word for a boy is 'Chello' in Konkani and 'Chele' in Bengali. The word for girl is 'Chelli' in Konkani, but Bengali uses 'Meye' for the girl. Some connection may be there with the English word boy and, the Sanskrit word 'Bālaka', Lativian 'Puika', and Lithuanian 'Berniukas'.

Finally, we take up Sanskrit forms and Dravidian Forms which were worked on by Aiyar. Figure 17 illustrates the MFAs for Sanskrit words and their Tadbhava forms in Drāvidian Languages.

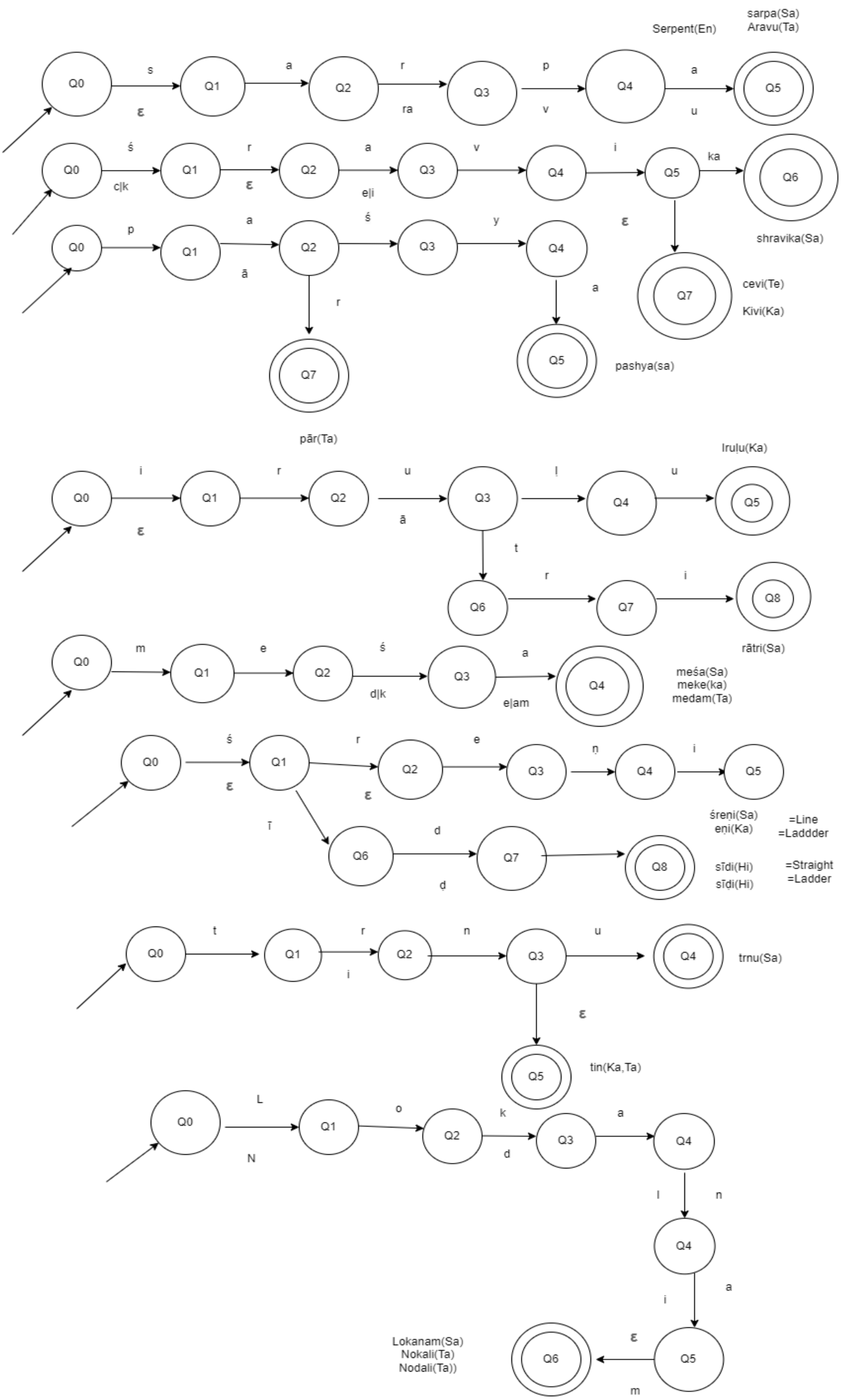

Figure 17 MFA for Sanskrit words and their Tadbhava Forms

In the first example, from the 'Sarpa' Sanskrit word first syllable is elided and the sound shift between pa and va sounds results in the 'Aravu', Tamil form which includes the suffix. The second example alludes to common origin for the word for ear in Sanskrit and Dravidian Languages. In the third case, 'Pashya' the word for seeing, is close to the Tamil form. In a similar vein, common words for night, sheep, night, and perceiving also seem to have commonalities. In summary, Finite State Machines serve as useful mechanisms for linguistic analysis across languages and can throw up not-so-obvious inter-relationships.

# 4. Discussions

In this section, we revisit the antiquity of Vedic Sanskrit, then reexamine how the languages are formed, in particular Sanskrit. This is followed by an analysis of word formation. Drawing on these analyses, we propose an Ecosystem Model for Linguistic Development with Sanskrit at the core.

## 4.1 Revisiting the Antiquity of Vedic Sanskrit

The speculated date of 1200-1500 BCE for Vedas opened the possibility of other Indo-European languages older than Vedic Sanskrit. These also led to the inference that the Indus civilization pre-dated the Vedas. Both these inferences are now widely questioned by scholars from fields as diverse as archeology to astronomy.

Amitabha Ghosh [32] analyzed the astronomical observations referred to in Vedic texts such as stellar conjunctions, eclipses, equinoxes, solstices as well as exaltation of planets such as Mars. These observations were picked from Vedas, Brahmanas, and associated literature and the plausible dates are arrived at using modern astronomical software. Table 26 lists the observations.

Table 26: Astronomically Derived Dates in Vedas

| Period | Dates of Astronomical Observations |
|---|---|
| Pre-Vedic and Early-Vedic | 8326 BCE<br>4677 BCE<br>4539 BCE<br>4350 BCE<br>4105 BCE |
| Vedic | 3961 BCE<br>3928 BCE<br>3541 BCE<br>3281 BCE<br>2948 BCE<br>2924 BCE |

These dates give very early provenance to Vedic Sanskrit. Vedic Scholar K. Suresh [33] divides the period of Vedic literature fourfold: (1) The period of creation of Mantras (2) The period of collection of Mantras (3) The Brahmanas period and (4) the Sutras Period. The seers involved were numerous spanning multiple generations generating mammoth literature. Rigveda had 21 Shākhās(branches/variants), Yajurveda had 109 Shākhās, Samaveda had 1000 Shākhās and Atharvaveda had 50 Shākhās There are references to 109 Upanishads. All this adds credence to the inference that Vedic literature would have taken millennia to evolve, and Vedic Sanskrit indeed is ancient.

Over and above this, Rigveda describes the river Sarasvati as fully flowing and merging into the sea. Now it is known that the river Sarasvati dried up by 1900 BCE or so. The dating of Vedas must be much earlier. There were questions about the lack of horse bones and chariots in archeological findings. These have been put to rest with recent findings such as at Sinauli. As Rigveda did not seem to have mentioned Iron, the date of use of Iron was another marker. Even that date, with new archeological findings, has been shifted back from 1200 BCE to 2000 BCE.

The Indus Culture and Vedic Culture were contemporary if not the same, as Vedic Altars have been found in many Indus Sites. Prof. Gaya Charan Tripathi [34] makes a plea to revisit the dating of Rigveda and covers these points eloquently.

## 4.2 How Languages Are Formed

On one hand, the European Scholars have hypothesized a Proto-Indo-European Language as the mother of Sanskrit and other Indo-European Languages. On the other hand, traditional wisdom in India considers Sanskrit as the mother of all Indian Languages. Dattaraj Deshpande [35] examines this quandary, with a unique perspective.

In his exposition, Deshpande first lists the hypotheses used in the field of comparative linguistics.

- H1: Every Language has a start date before which it did not exist.
- H2: Languages are pure in their original form and then get corrupted or decay over time
- H3: Languages change with locality and over time
- H4: Languages loan words to one another
- H5: Words get modified beyond recognition due to faulty pronunciation
- H6: Languages that have the most distorted words are older. Older languages are simpler and raw.
- H7: Original language has pure and precise pronunciation. Borrowing language has distorted and corrupted pronunciation.
- H8: A linear sense of time of Western scholars, in place of cyclical as in Indian Tradition
- H9: With time the world gets more and more chaotic. (2$^{nd}$ law of thermodynamics)

Languages go together with cultures and civilizations. The way words are pronounced changes with date, time, climatic conditions as well as the ability to pronounce them. A child, a person with speaking disabilities will invariably distort words. Another aspect is the effort to pronounce a word, in colloquial settings, the words tend to get simplified, and simpler forms are more popular. Different regions as well as languages prefer certain sets of sounds.

In the above list, hypotheses H6 and H7 contradict each other. But both phenomena are possible. According to Western Linguists, the original languages are more primitive. Secondly, natural languages decay over time and based on such decay the age of language can be assessed. Both these assumptions do not hold for Sanskrit. With Sanskrit there has been a constant focus on precision, Sanskrit has generally co-existed with Prākrit's or colloquial forms where such decay is possible. But that decay is continually arrested when the language is used in say Vedas, as there is a strong focus on preserving the hymns without any error. Secondly, Sanskrit preserves multiple forms of the same word and routinely reintroduces the formal word back into Prākrit's or natural language. There was so much stress on preserving exact pronunciation that the sounds and words are retained even when the meanings are a matter of debate. Ram Gopal [36] details multiple ways Vedas are interpreted. That debate has gone on for centuries. Deshpande considers one language being the mother of another language as not the right way to look at linguistic development. He seems to allude to the fact that languages co-evolve over time. Thus, he considers Sanskrit being the mother of all Indian languages as not the right way to frame the question.

Sanskrit is endowed with a rich and robust grammatical tradition that allows for auto-correction, and a community of scholars takes responsibility for that. In addition, multiple ways of memorizing Vedas from Padapāṭa to Ganapāṭa provided a mechanism that is even more powerful than modern checksum, to ensure the integrity of the text. Even the pitches, accents, and tones were preserved for generations. Because of such precision and richness, Buddhist literature which was initially in Pali moved to Sanskrit.

## 4.3 How Words Are Formed

There probably is a huge body of literature in linguistics on word formation. However, there are probably rather few studies with a perspective that is as grounded as it is exalted as done by Ram Swarup.

Ram Swarup [37] does a detailed analysis of how the words are formed. He starts with the observation that certain sounds singly or in combination express certain phenomena or emotions. Among the Pāninian sounds, some express softer sentiments whereas others are virile. Then he dwells on how things are named. According to him, many new things are named based on the names of older things unless they are completely new. Then he looks at the roots of words. Sanskrit is very rich in roots. Then he describes synonyms and how multiple words are used for the same phenomenon such as fire using different roots that manifest different aspects of fire - lighting, purification, etc. Then there are 'manas' words and 'buddhi' words. Here former connects with mind/sense perception and the latter with intellect/cognition enabling abstract concepts. Then there is a continual attempt to unlock higher meanings into words. In the words of Ram Swarup: 'Word is a living thing. It is pregnant with life and possibilities. It grows and expands meanings in a hundred directions. The process of unfoldment and development, like all truly vital processes, is unconscious, but truly intelligent and wise'.

In general, Sanskrit has served as a morphological foundry for many languages. Thus, Sanskrit lives through words in other languages. For example, to craft a word to mean empowerment, a modern concept, in Sanskrit as 'Sabalikarana', is rather effortless. Thus, Sanskrit continues to be the destination of new concepts and words.

## 4.4 Ecosystem Model for Linguistic Development

Our analysis so far drives us to the conclusion that the formation of words and languages is way too complex to be explained by a family-tree model of languages. An alternative model is required that is more holistic and harmonizing.

To that end, we propose the Ecosystem Model for Linguistic Development. Here Sanskrit (Vedic or otherwise) invariably develops in intimate proximity with natural language/Prākrits where the speakers of natural languages contribute to Sanskrit and Sanskrit in turn enriches natural languages, by being a donor for words. Then over time certain languages and words migrate, in other cases, Sanskrit itself migrates either as a language, literature, or technical knowledge. In some cases, the words migrate as is whereas in other cases words change beyond recognition and only with a good degree of analysis, the common roots and basic words can be discovered. With every language in currency, certain sounds may be preferred and certain other sounds rarely used. Some sounds may be unique to a language. All these lead to the adaptation of words to a new milieu. Then these words and expressions are reorganized as per the evolving grammar of languages. Figure 17 depicts a Linguistic Ecosystem with Sanskrit at its core. Here Prākrits are literary languages that have served as an alternative to Sanskrit, in contrast to Bhasha or vernaculars widely used as colloquial languages.

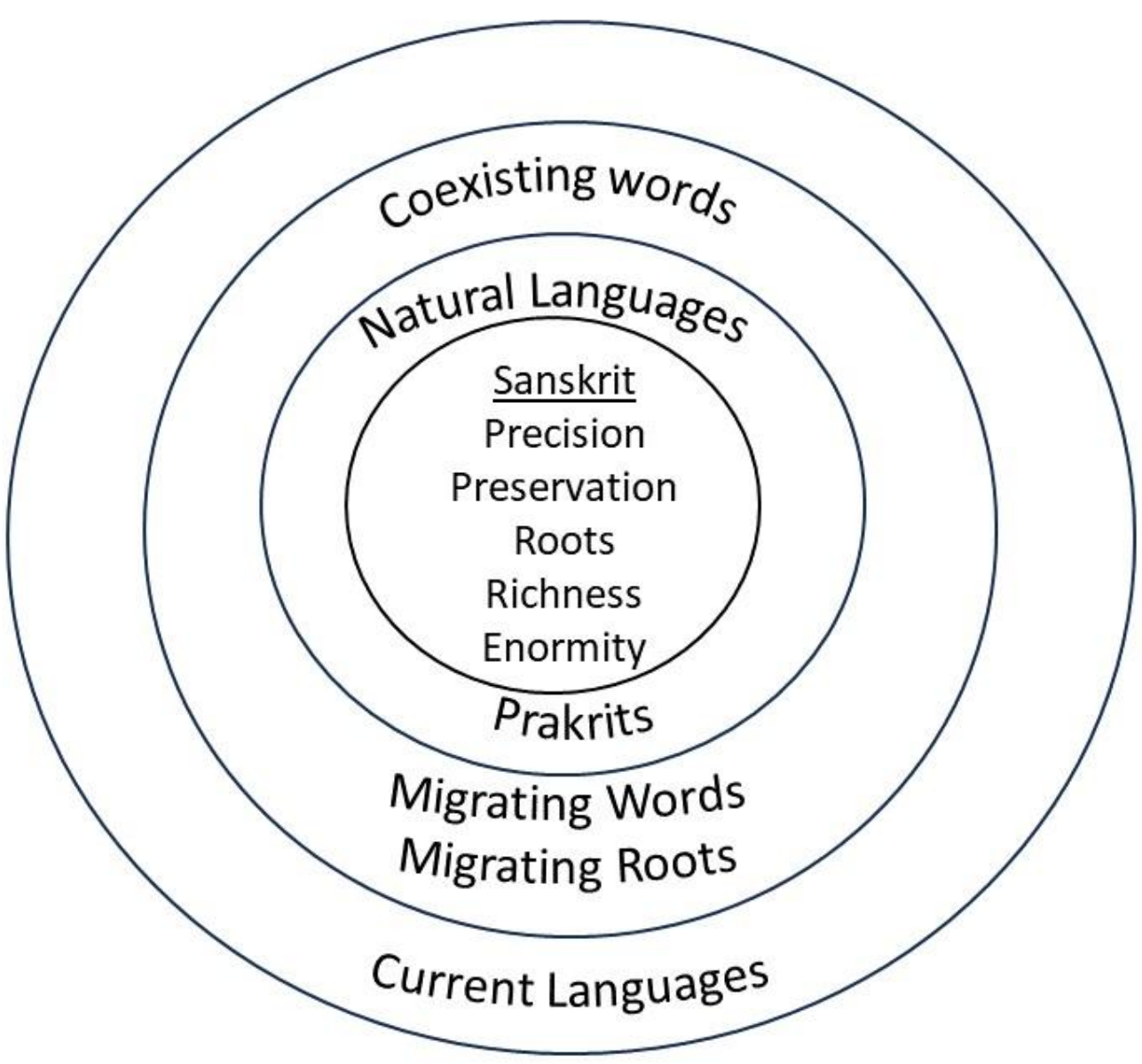

Figure 18: Linguistic Ecosystem

Figure 19 illustrates the ecosystem phenomena at play.

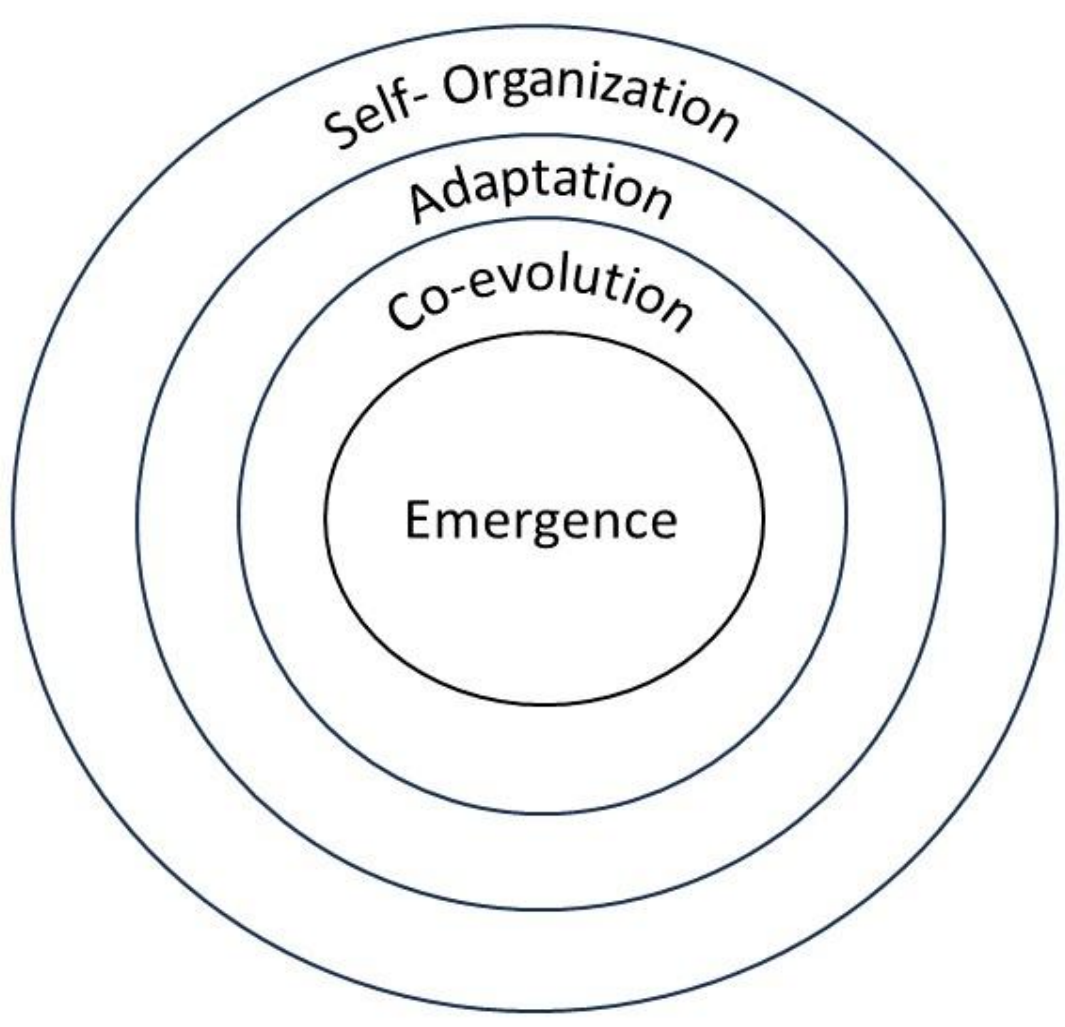

Figure 19: Ecosystem Phenomena at Play

Firstly, the formation of words can be described as an 'emergence' phenomenon. Here players participate without necessarily being conscious and purposeful. Thus, certain sounds get associated with certain meanings and get used to refer to certain entities. Then the existing words in part or whole contribute to the formation of new words. Words may have a particular meaning at one time and a radically new meaning may emerge for the same word due to social phenomena. Secondly, the words, as well as languages, 'coevolve' due to varied interactions, commonality of speakers, shared spaces and times, beliefs, traditions, culture, and civilization. Thirdly, as words move from language to language and language moves from region to region they get exposed to new climatic conditions and new groups of speakers. Such movements will result in the adaptation of words and languages. The context for the use of language is another important dimension that determines the nature of adaptation. In certain contexts, precision gives way to ease of use. In other situations, the sounds may get shifted to the extent that it becomes rather hard to link the changed word with the original word. Finally, every language has a certain self-organization which sets rules for word and sentence formation. This organization may be cognitive based on well-defined grammar or habitual based on collective behaviors. Even here languages may be influenced by neighboring languages. For example, Indian Languages commonly follow the subject-object-verb order unlike English which follows the subject-verb-object order. Here Sanskrit is unique where the meaning does not change with the order of parts of speech.

The ecosystem is a very powerful paradigm. The paper [38] dwells on the ecosystem paradigm extensively as it applies to the complex agricultural sector.

## 5. In Retrospect

Any observation has four facets: the observation itself, the object of observation, the context, and the observer. Here the observer has disproportionate influence and so does the context. This particularly applies to comparative linguistics or social linguistics research. Here, certain researchers and certain inferences are considered mainstream, and research work in line with their research work gets prominence and the rest, often tends to get ignored. For instance, the seminal work on Dravidian Languages by Swaminatha Aiyar [25], which explores the intimate connection between Vedic Sanskrit and Dravidian Languages is seldom cited in the literature. Further, the position of G.U. Pope who was contemporary to Caldwell [39] on Dravidian Languages is relatively unknown even though Pope authored important scholarly works on Tamil [40,41]. In this section, we delve into this contestation of ideas and cover counterviews and alternate perspectives.

The context of observation is equally important. For centuries, the Aryan Invasion Theory was considered a given, leading many scholars to retrofit their findings into that framework. Even when they found many issues they tried to look for an alternate explanation.

With the above backdrop, we provide a linguist perspective, a language enthusiast perspective, and an immersive perspective, followed by a philosophical perspective, with the commonality that each is holistic.

Coming to linguistics, there is a dominant view of Western scholars, shaped by what they have observed elsewhere. This causes blinders when they analyze India, which has very different philosophical underpinnings from the rest of the world. Their position is that when two languages come into contact one language is elite, another is subservient, and the speakers of the subservient language are forced to adopt the elite language. This is called the substratum theory. The corollary of substratum theory is that the Dravidian language speakers while shifting to prestigious Sanskrit, changed Sanskrit by imparting features of Dravidian Languages, which are absent in European Languages. Certain scholars who cannot make a case for Dravidian influence, attempt to name it Munda influence. This helps to establish that Sanskrit is an import into India and that the speakers of Dravidian and Aryan languages were different people. This view is referred to as the migration view. However, India is known for polyglossia, polygraphia, and mutual coexistence/respect of languages. Till the colonial period, languages never caused divisions. Rather, people took pride in being multilingual and the Aryan-Dravidian binary was unknown to Indians till the 1800s.

In this context, the perspective argued by Sonal Kulkarni Joshi [42,43,44] is significant. Kulkarni-Joshi [42] contests the Dravidian substratum in Marathi proposed by Southworth [45]. She proposes contact theory and the relatively late influence of Dravidian Languages on Marathi. She traces the provenance of linguistic features across Maharashtri Prākrit, Old Marathi, and contemporary Marathi. Southworth's work primarily compared contemporary Marathi and Dravidian languages. According to her commonality can be explained by borrowing and diglossia. Her examination suggests the absence or rare presence of some key morpho-syntactic structures in Proto- and Old Marathi. She also contests that the words considered part of Dravidian core vocabulary are necessarily Dravidian. They could as well have been Indo-Aryan. There is also the near absence of structural influence of Kannada or Telugu on Marathi.

In her next paper, Kulkarni-Joshi [43] critically examines the substratum view in the context of the Aryan Migration view and the Out-of-India hypothesis. She reiterates the view of Western scholars that "the languages of India, even though divided into language families were considered part of a single linguistic area [46]. The shared features include retroflex sounds, Subject-Object-Verb order in sentences, absence of prepositions, morphological reduplication, echo formation, reduplicated verb adverbs, explicator compound verbs, use of converbs, oblque marked subjects and morphological causatives among many others".

Kuiper [47] identified 4% of words in Vedic Sanskrit as without equivalent Indo-European vocabulary. The loan words seem to have increased in later Mandalas of Rigveda and Classical Sanskrit then declined in middle Indo-Aryan and modern Indo-Aryan periods

The Substratum Theory was severely critiqued by Trautman [48,49]. Hock [50,51] did not accept that retroflex consonants were a Dravidian influence on Sanskrit. Rather, he claimed that it was internal innovation. Kuiper [47] and Emaneau [46] consider Retroflexes as a parallel development between Indo-Aryan and Dravidian. Hock [50] considers retroflexion as pre-vedic. He also points to structural differences in the way retroflection developed in Indo-Aryan and Dravidian languages. In any case, retroflection is not unique to Sanskrit, it is very much present in Norwegian and Swedish languages. Diglossia and polyglossia prevailed in India and people learned languages based on their needs and not out of compulsion. Kulkarni, concludes leaning in favor of the migration view and does not accept Sanskrit as PIE. She says the Out of India hypothesis does not appear to be based on linguistic evidence and is difficult to verify. However, she did not counter Mishra's evidence [19] in a clear-cut manner and does not dwell adequately on Talageri's work.

Talageri [52-54] provides clear-cut arguments in favour of the Out-of-India hypothesis and against the Aryan Invasion Theory. Talageri claims since the linguistic evidence is not conclusive to establish the Aryan Invasion theory, the scholars turned to Rigveda to look for clues of invasion. He disputes the Aryan Invasion theory in great detail and argues that Rigveda is very much native to India and the movement of Rig-Vedic people has happened from the East(Western UP and Haryana) towards the Northwest, most references to the Afghan region are in later Mandalas. Chronological Analysis [55] of deity and river mentions using clustering and social network analysis corroborates common, continuing civilization on one hand and steady geographic movement towards the northwest on the other, of Vedic composers. Further, Talageri establishes that the movement of historical narrative in Rigveda is also from east to west. In addition, archaeologists have not seen any discontinuities in material culture in Vedic region between 1000 BCE to 4000 BCE. On the other hand, there is archaeological evidence in Europe where IE speakers migrated.

Talageri is a strong proponent of the Out-of-India Theory and lays out his argument thus. There are twelve known branches of Indo-European Languages: Italic, Celtic, Germanic, Baltic, Slavic, Illyrian (Albanian), Greek, Thraco-Phyrgian (Armenian), Hellenic (Greek), Anatolian (Hittite), Iranian, Tocharian and Indo-Aryan. The isoglosses shared by different branches of Indo-European can show the order of migration of branches from the presumed homeland. Except for Hittite, the other 11 branches share many basic linguistic features. That indicates that Hittite was the first to migrate/separate from the homeland. Five branches (Indo-Aryan, Albanian, Armenian, Greek. and Iranian) share certain late features missing in the other branches, which likely developed after the migration of seven other branches.

Linguists except Mishra do not accept that Sanskrit is archaic enough to be the mother of all languages. Thus, they hypothesize PIE language as a common ancestor. Then they look for the homeland for PIE. Hock [56] proposes a model where the relative geographical position of languages in the homeland is the same as the current configuration. Thus, he explains that the Indo-European Language branches moved from somewhere in Russia to Europe and maintained the same relative position. However, he is not able to account for Indo-Iranian languages which moved eastwards. Then he leaves out the Tocharian language from his fan-out model. He maintains that the scenario of the Indian Homeland for PIE and languages trickling out through the Northwest is harder to accept.

Talageri points out that Tocharian, Hittite, and Italic shared important isoglosses and possibly were proximally located in the homeland. In the current positions however, they are located in different corners and do not corroborate the fan-out model.

According to Talageri's analysis, Hittite, Tocharian, and Italic share few unique isoglosses and were the first, second, and third to migrate out. Talageri considers India (Haryana to Afghanistan) as the primary homeland and these dialects moved to a secondary homeland (Central Asia). Indian historical tradition records the movement of the Druhyu and Anu tribes westward.

Gamkrelidze [57] postulates two major dialect areas: Area A comprising Anatolian-Tocharian-Italian-Celtic, and Area B comprising Indo-Iranian-Greek-Balto-Slavic-Germanic in the homeland area. These two dialect areas functioned independently and had distinct structural innovations. Within area B, there were two distinct sub-areas, B1: Indo-Iranian-Greek-Armenian and B2: Balto-Slavic-Germanic. Talageri infers that only the Indian homeland can explain such geographic dispersion, wherein Area B is the original homeland (Haryana to Afghanistan) and Area A is the secondary homeland (Central Asia). He elaborates on six stages the languages exited from their primary homeland to their secondary homeland and then to their current location, by making use of isoglosses that buttresses the hypothesis. The complete absence of isoglosses between the first two branches that exited (Hittite and Tocharian) and the last two branches remaining (Indo-Aryan and Iranian) corroborates the hypothesis. He then analyses how the Semitic word wine was borrowed in the 9 Indo-European branches that went westwards but not in 3 branches that remained in the east, namely Tocharian, Indo-Aryan, and Iranian. Talageri counts 3 migrations. Hittite migrated to Central Asia then around the Capsian Sea migrated to Anatolia (Turkey). Five European branches (Italic, Celtic, Germanic, Balto-Slavic) went to Central Asia, Siberia, steppes and then to Europe. The last 3 branches passed from Iran to West Asia were Greek, Albanian, and Iranian. The Armenian language remained somewhere in the center.

Whereas Talageri is open to the PIE hypothesis, he is particular that is native to India. He studied number systems [58] used in Indo-European and other languages and divided them into stages 1-4, based on complexity. He considers, Sanskrit, Tocharian, and spoken Simhala to belong to Stage 2. The 9 IE branches that left India belong to stage 3. So do Dravidian Languages. Only contemporary Northern Indo-Aryan Languages belong to stage 4, where they have the maximum number of unique expressions for numbers 0 to 99. This further indicates the archaic nature of Sanskrit and the continuing evolution of languages in India. Further, the Simhala language is close to the Northwestern language and retains archaic words, for water, grass, etc. The language is located beyond the Dravidian region. This also attests to the possibility of Indian origin of the Indo-European Languages.

In the third paper [44], Kulkarni argues that the European belief system which prized unity and homogeneity of languages was at variance with the Indian system which prided multiple languages that blended well. In the Indian system, sound or nāda was a core construct. In contrast, the Western system had words and reason(logos). She refers to the 18th-century British perspective and then the 19th-century North American perspective. She says even the notion of languages i.e. Bengali or Assamese being distinct from each other is a British imposition. Otherwise, India just had a sea of dialects each claiming to be language on its own yet part of a linguistic continuum within a geographic continuum. Codification of languages and identification of mother tongues happened side by side Linguistic survey of Greison that took place between 1903 and 1928. The scripts were also multiple for the same language. Some scripts such as Modi script (for Marathi) were done away with in the process of standardization. All these made language a divisive force. Slowly the languages were used to divide people using the ethology-philology nexus. They divided the people while uniting languages into different families.

In the 1900s, North Americans funded social linguistics studies via linguistics departments set up in many Indian universities. Here American and Indian scholars worked side by side. In their social-linguistic studies, American scholars emphasized differences and overlooked commonalities. They took an anthropological perspective which did not suit India. They looked at everything with the prism of caste. "In the process, the vastly complex and often fluid features of Indian reality were oversimplified, misinterpreted, and misrepresented. Indians were assigned to stereotypical boxes such as caste, religion, mother tongue, vernacular, language, and so on. Table 27, traces the journey of the formation of modern Indian languages.

Table 27: Formation of Modern Indian Languages

| Language | Branch |
|---|---|
| Vedic Sanskrit | Prākrit, Classical Sanskrit |
| Prākrit | Elu, Magadhi, Pali, Maharashtri, Shauraseni, Gandhari |
| Elu | Dhivehi, Vedda, Sinhala |
| Maharashtri | Konkani, Marathi |
| Magadhi | Bihari (Bhojpuri, Maithili, Magahi), Odiya, Bengali, Assamese |
| Shauraseni | North-Indic (Dogri, Punjabi, Sindhi)<br>West-Indic (Marwari, Romani, Gujarati)<br>Dardic (Kashmiri, Shina)<br>Pahari (Nepali, Garhwali, Kumaoni)<br>Hindustani (Hindi(Haryanvi), Urdu(Rekhta, Dakhni)) |

The above grouping however has many caveats and assumptions. For example, Konkani and Marathi are shown close to Maharashtri Prākrit. However, Konkani has distinct features such as nasal vowels that are common to Hindi and many features common with Northern Languages and it may have different origins [28]. It is also interesting to note that Konkani, being primarily a vocal language has retained the original sounds and constructs with greater authenticity than it would have if it were to be a written language,

According to Aiyar and Mishra, even Dravidian Languages will fit into the same framework as far as bulk of words and linguistic features are concerned, allowing for a degree of native words and characteristics. Sanskrit has aided in mutual intelligibility between Dravidian

Languages [59]. What is a language in its own right and what is a dialect is also a matter of contention. Many dialects are grouped under the Hindi language now [60]. Odiya once labelled as a dialect of Bengali is considered a language now. Figure 20 below shows retroflexes were very much part of Indo-Aryan Languages and not limited to Dravidian Languages. The figure shows the presence of the retroflex l sound in Western dialects and its absence in Eastern dialects.

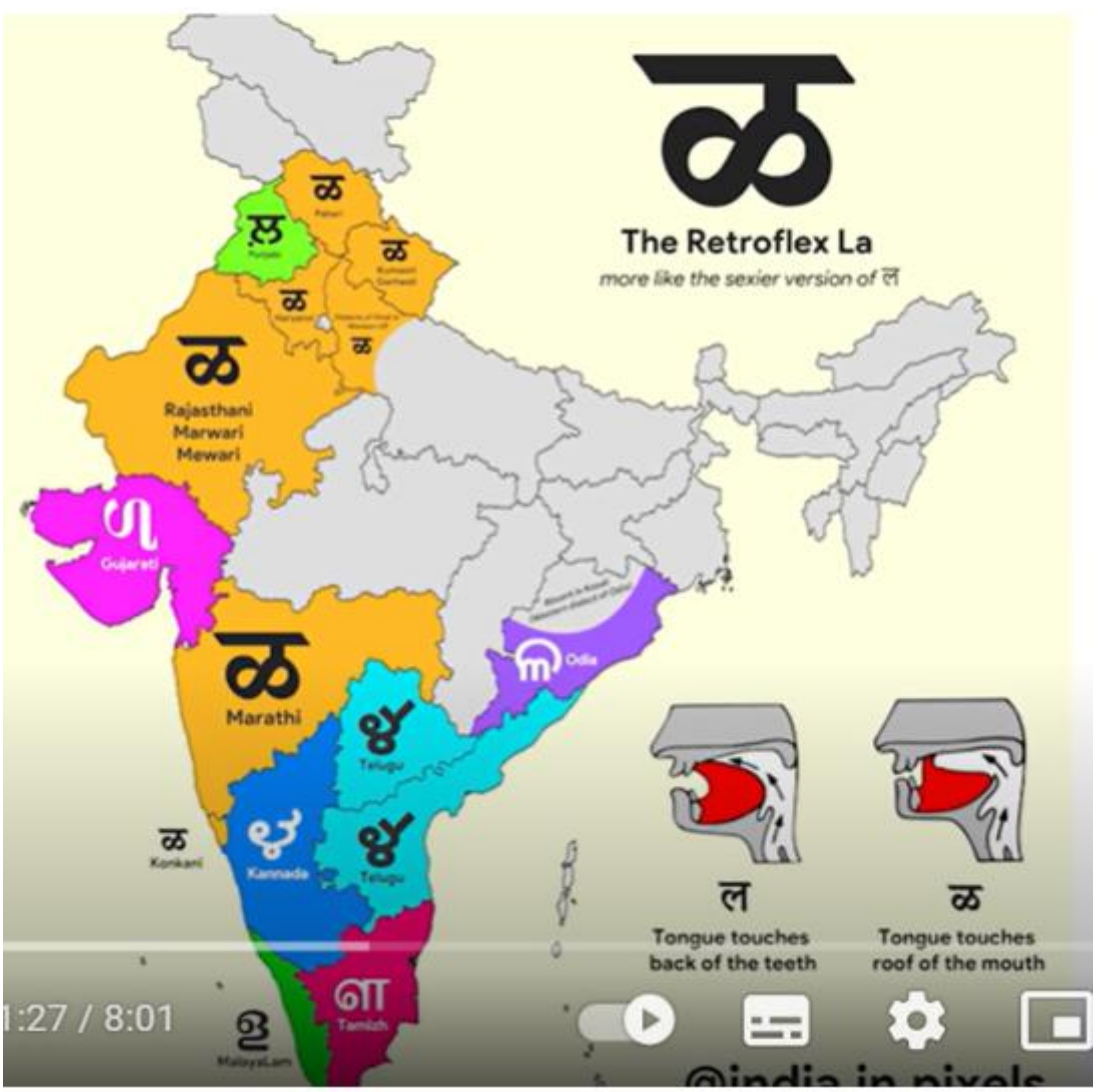

Figure 20 Retroflex La sound in Indian Languages (Credits: India in pixels)

The Indian languages differ based on their preferred sounds and vowel usage. In the Eastern direction Magadhi Prākrit, Sanskrit became softer, “rounder”, and rhythmical. In the north side, Sanskrit became coarser, firmer, and more masculine. Towards the South, Sanskrit became more dramatic and became Maharashtri Prākrit,

Marathi and Kannada have features that are not common in other Dravidian languages. So do Telugu and Marathi. Each language has a unique personality. Marathi and Telugu have unique retroflexes and alveolar sounds. Telugu tends to be musical and Bengali sweet and aesthetic, Telugu and Bengali maintain vowel harmony. Ashris Chaudhari [61-63] has shared lot of interesting nuances about Indian Languages. The way in India Prākrits gave rise to modern language, in Europe Vulgar Latin gave rise to modern languages such as Italian, Spanish, and French.

Shulman in his "Biography of Tamil" [64], studies Tamil poetry and literature from Sangram era of ancient times to modern times, in an immersive manner, and provides valuable insights as a poet and scholar. He covers the periods of Pandyas, Cheras, Cholas and Satyaputras in detail. He refers to role of Agastya, who is credited as originator of Tamil grammar and culture Tolkappiar as well to Dandin a Sanskrit poet of Tamil region. He discounts the North-South or Aryan-Dravidian divide. According to him Tamil has co-existed with Sanskrit from the very beginning in an environment of diglossia, where Sanskrit is used for ritual purposes and formal settings and Tamil in colloquial settings. Over time, however Tamil like other South Indian Languages is replete with Sanskrit words. However, Sanskrit has retained purity with minimal substratum effect, despite co-existing with South Indian languages for millennia. Any claim about a distinct Tamil civilization is a totally modern construction which has no bearing on historical reality. Many Western scholars see everything using the prism of caste and only caste. Shulman stands apart from them.

All Tamil Kings from Pandyas to Cholas were patrons of both Sanskrit and Tamil and they saw a symbiotic role between them. Mani Pravalam where Sanskrit and Tamil words become part of a literary formation the way rubies and corals, where unity is demonstrated by colour despite intrinsic value addition.

Andrew Ollett[65] studied language order in India, among Sanskrit, Prākrits, and Bhasha(vernaculars).. Shulman observes that Andrew Olett's perspective on the inter-relationship between Sanskrit and Prākrits holds for the South Indian Language system as well. "Tamil and Sanskrit constitute a set in which each language intrinsically defines the other. Each both contrasting with and complementing the other. The terms included In this dyadic set are mutually constrictive" Fault lines and divisions came only after colonial missionaries and caste-based politicians gained influence in the last two hundred years. Unfortunately, the divisions are continuing.

Shulman while analyzing relationship between Tamil and Sanskrit refers to the work of Alexander Lubotsky [66], Kuiper, and Witzel. Here Kuiper saw Dravidian substratum in Vedic Sanskrit, whereas Witzel identified Munda words. Lubotsky in his paper on Indo-Iranian substratum hypothesized the substratum effect outside India. In his words, "The phonological and morphological features of Indo-Iranian loanwords are strikingly similar to those which are characteristic of Sanskrit 1oanwords, i.e. words which are only attested in Sanskrit and which must have entered the language after the Indo-Aryans had crossed Hindukush". A simpler explanation however would be that Iranian and Indo-Aryan languages co-existed in the region between Punjab to Afghanistan and remained there long after European branches left, during which they developed words and features unique only to Indo-Iranian languages.

Sanskrit is a rich source of words for languages of the world, a morphological foundry. In English, 90,196 lexical entries lead to 317,477 forms, with a ratio of 3.5:1. In Sanskrit, a 170,000-strong lexicon leads to 11 million forms, with a ratio of 64:1. This unique capability of Sanskrit was lauded by Frederik von Schlegel [67] in his treatise that covered languages, cultures, philosophies, arts and aesthetics, and history.

To quote Schlegel

- ’The Indian grammar harmonizes so completely with the Greek and Latin, that it appears to be scarcely less closely connected with those languages than they themselves are with each other. The similarity of principle is a most decisive point; every degree of modification or comparison being expressed, not by the addition of separate words, whether particles or auxiliaries, but by inflexions, throughout which the peculiar form of the root is distinctly preserved”.
- “In the Indian and Greek languages each root is actually that which bears the signification, and thus seems like a living and productive germ, every modification of circumstance or degree being produced by internal changes; freer scope is thus given to its development, and its rich productiveness is in truth almost illimitable. Still, all words thus proceeding from the roots bear the stamp of affinity, all being connected in their simultaneous growth and development by community of origin. From this construction a language derives richness and fertility on the one hand, and on the other strength and durability. It may well be said, that highly organised even in its origin, it soon becomes woven into a fine artistic tissue, which may be unravelled even after the lapse of centuries and afford a clue by which to trace the connexion of languages dependent on it, and although scattered throughout every part of the world, to follow them back to their simple primitive source. Those languages, on the contrary, in which the declensions are formed by supplementary particles, instead of inflections of the root, have no such bond of union, their roots present us with no living productive germ, but seem like an agglomeration of atoms, easily dispersed and scattered by every casual breath.”
- ” All the preceding proofs appear clearly to establish the fact that the Sanscrit or Indian language is of higher antiquity than the Greek or Latin, not to mention the German and Persian. We might, perhaps, decide more satisfactorily in what relation it stands, as the earliest derived language, to the general source ; if it were in our power to consult the Veda in its genuine form, together with the vocabularies which were early required on account of the great difference between the language of the Veda and the Sanscrit”
- ” The Saga of Ráma, who is described as a conqueror of the wild tribes of the South, might seem to favour the opinion that the Indian language, even at a very early period, suffered considerable foreign intermixture from the various tribes incorporated with the body of the nation. The northern part of the country is peculiarly the seat of the Indian language and philosophy. In Ceylon we still trace the influence of the foreign tribes of Singhalese, which in former times was probably of more extensive operation. Still the regular, simple structure of the Indian language proves that the in fluence of foreign intermixture was never so overpowering or heterogeneous as in other languages of the same family.
- “It would, perhaps, be too much to assert without reservation that the Greek and Latin languages hold the same position in regard to the Indian as the Italian does to the Latin, although it is undeniably true that a certain irregularity of form, and the use of prepositions in those languages, already presage the transition to modern grammatical construction; and the regular simplicity of the Indian language in parallel cases is an incontrovertible evidence of greater antiquity.”
- **“** It would be difficult to point out any idea or doctrine, common in either of the different intellectual systems, which was not also known among the Indians ; nor any fable holding a distinguished place in merely poetical mythologies, the counterpart of which does not exist also in the Indian.**”**

- “It must not be forgotten that India has always been one of the most populous countries in the world, and is so even at present, notwithstanding the numerous destructive revolutions of the last century and the prevalence of universal misery and oppression. How natural, then, is the inference that the overflowing population may have rendered emigration a measure of absolute necessity at the period of its ancient prosperity? It may not necessarily be invaders. It may be priests, royalty, or anybody. The migrations within India are a point to note.”

Schlegel did a comparative study of words in Sanskrit, Greek, and Latin. The words he referred to in his work are reproduced in Annexure 4[68]. Schlegel who took a multi-disciplinary perspective that spanned language, culture, religion, and philosophy was certain that Indian civilization was the seed from which all others sprouted. His views and profound insights are worth revisiting by contemporary scholars.

# 6. Our Contribution

In this section, we contextualize our work in fields such as Computational Morphology, Machine Learning, and Vector Databases, distinguish it from conventional approaches taken hitherto, and highlight future research avenues.

The use of the state machine model in natural language processing is not new. But the majority of applications view each word as a token and do the analysis at the level of phrase or sentence and typically for a specific language. In this work, we make use of the state machine model to study word formation across related languages. In particular, inflectional languages transform roots in myriad ways to create new words. In addition, when a word moves from a language with one set of sounds what we refer to as a morphological alphabet to another with a different morphological alphabet, it changes at times radically. Then the vowels get manipulated for languages to sound musical, sweet, and easy to express, etc. All these can be analyzed using the state machine model. It is also possible to construct a finite state transducer to migrate a word from one language to another so that it blends in with the target language.

In addition to the state machine model which belongs to discrete space, we have modeled words using continuous vector space. Each human sound is given a unique coordinate based on Panini’s system of sounds. Then we represent words as paths on the phonetic map. In this process, each word is represented as a phonic signal on a domain of sounds that has a specific geometry. This is very different from representing word as a sequence of sounds collected from a bag of sounds and ordered in different ways. The phonetic map proposed in this paper is useful for generating a ‘Universal Lexicon’ catering to a plurality of languages. A patent application [69] is filed in this regard. Further, once every word is represented as a vector, emerging Vector databases can be used to store and analyze them.

As each word is represented as a vector of sounds, we can train machine learning models on words of the source language and destination language, framing the problem as regression or classification, as the case may be. Further research in this area can open up avenues in natural language translation and generation.

## 7. Vocabulary

The words we have used for our analysis and inferences are given in the following annexures.

Annexure 1[31], provides a list of cognate and related works across Indian and European Languages, based on our analysis.

Annexure 2[26], provides a list of words sourced from Dravidian Theories [20] which demonstrate the linkage between Sanskrit and Dravidian Languages.

Annexure 3 at the end of this paper, provides further details on word formation and can give insight into the diffusion of roots, words, and meanings across the linguistic ecosystem. Here you can see that commonality of words cuts across seemingly unrelated languages and from a single root a wide canvas of words is created that spans the whole linguistic ecosystem. Here we have sourced some of the words from Ram Swarup's work [37].

Annexure 4[68] provides a list of words, sourced from Schlegel [67] where he majorly compared Sanskrit, Latin, and Greek words.

Annexure 5, a list of miscellaneous words across languages is reproduced at the end of the paper,

## 8. Conclusions

In this paper, we have analyzed languages with a particular focus on words. The words are divided into word groups where a set of these words form m-language (morphological language). With a given m-language, we associate an m-alphabet. The m-alphabet may have a basic version with common sounds and an extended version with all sounds. Corresponding to these morphology-based constructs we construct state transition diagrams, here every phoneme is a state and so is a sequence of phonemes. A valid word, a member of m-language is an accepting state. A suitable grammar can thus determine whether a word belongs to the word group or not. To enable that we construct a unified Morphological Finite Automata which is expressed compactly and accepts all words belonging to the m-language, that cuts across multiple natural languages. Secondly, this exercise can enable us to infer new words that may belong to the same word group and give insights into hitherto unknown associations between two words either belonging to the same or different languages.

We have used Pānini's System of Sounds to represent sounds and words. In addition, we have defined a phonetic map that geometrically manifests these sounds on a 2-dimensional plane. Thus, each phoneme has a coordinate on the phonetic map. Each word has an associated distance measure that indicates the quantum of traversal required on the phonetic map. This measure we have used to analyze differences between words. Thus, based on the distance we can term some words as basic words, some as refined words, and some others as central words. These ideas we believe are useful in comparative linguistics.

The phonetic-map distance measure we believe is an improvement on the current mechanism to compare words in natural languages. One approach is to use Levenshtein Distance, where natural language words need to be transliterated first in English. Here the number of substitutions/modifications required to get two words to match is used as distance. This misses the phonetic dimension. The second well-known measure is Soundex. This works well for European Languages, in particular for de-duplication of names. Here each word is associated with a code such as M460. Soundex uses the following codes: 1=B,P,F,V; 2=C,S,G,J,K,Q,X,Z;3=D,T;4=L;5=M,N;6 = R. The letters A, E, I, O, U, Y, H, and W are not coded. Compared to these measures the scheme we have proposed is more elaborate and promising. In an earlier paper [70], Soundex-based measures were used for language classification.

Based on our analysis in this paper, we surmise the following: Vedic Sanskrit as part of Chandas (prosody) has retained the most refined forms from which simpler forms can be derived. Thus, in certain cases, a word in Sanskrit may result in a high distance measure on the phonetic map. Also, the Sanskrit word in many cases is a central word that has cognates cutting across languages, and language groups. If we were to use a genetic or clustering viewpoint, Sanskrit words have some relationship or other in some manner/context or other with all other languages among the Indo-European Languages. At times Greek/some other language may appear to have a more basic or original word compared to Sanskrit, but when you do the same analysis at the word group level that includes derived and related words, Sanskrit words are indeed central. Secondly, Sanskrit is *th*e donor language when it comes to the Dravidian Languages, even for day-to-day words. Hence, based on morphological analysis, a more accurate representation for the comparative linguistics field may be Sanskrit occupying the hub from which words have been transmitted to all other languages and groups of languages that underwent transformations in transit. The process of transformation of Sanskrit words in Indian Languages and European Languages is similar. This process has very likely happened over millennia due to well-acknowledged migrations within India and less understood outward transmissions to Europe.

Further, based on the insights gained from this study and drawing on the wisdom of Sanskrit scholars rooted in Indian tradition, we propose an ecosystem model for the analysis of languages in place of the genealogical model. With the genealogical model, languages are born and die, giving rise to other languages in the interim. Languages age over time making them almost intelligible, if the speakers of distant generations were to converse. Along with this are tied the hypotheses that make certain sounds older and primitive and inferences are drawn giving or denying motherhood/ancestry to languages. With the ecosystem model, words and languages emerge due to complex interaction, orderly and refined forms are preserved and multiple forms of words co-exist. Then words and languages coevolve as they participate in a common civilization, culture, or context. Then as words and languages migrate, they adapt to newer geographies and preferences/limitations of the users of language. The context of use also guides these adaptations. Finally, rich and robust grammar can organize this evolution in a guided manner, which Sanskrit has been and continues to be particularly well-endowed with.

## Annexure 3: Word Formation

Generally, the first two letters are used to indicate language. The languages listed are supported in Google Translate (translate.google.com). Retroflex sounds are capitalized.

| Word(English) | Cognates (Tatsama/Tadbhava |
|---|---|
| Pāth | Path(Sa) Hādi(Ka) |
| People | Lok(Sa) Log(Hi) |
| Group | Group(En), Gumpu(Ka) |
| Cave | Gavi(Ka) , Cave(En) |
| Colour | Varna(Sa) BaNNa(Ka) |
| Clan, Parent's house for married lady | Kula(Sa), Kula(Ko) |
| Old Person | Vraddha(Sa) BooDa(Hi) |
| | Aristo(crat) (En) Shreshta(Sa) |
| Throwing and flying | UDai(Ko) Throw<br>UDana(Hi) Fly |
| Covered | Kavida (Ka) MoDa Kavida Covered with Clouds<br>Covered (En) |
| Namaste | Vande(Sa) VaNakkam(Ta) |
| | Soona (Hi):Depressing, Shoonya(Sa):Zero, empty |
| Medicine | Medicine (En) Maddu(Ka) |
| Search | Shodh(Sa) So~dhi(Ko) |
| Touch | Tvak(Sa), Touch(En) |
| Cover | Topar(Sp) Topi(Hi):Cap |
| Proof/Evidence | Purave(Ka) Proof (En) Prova(Po) Purava(Ma) |
| Native: Born in a place, belonging to a place. | Naadu(Ka) our land, Naadiga- Person belonging to Naadu, Naati(Ka) – may mean something else. |
| Money | Cash(En), Kaasu(Ka) |
| Money | Paisa(Hi), Peso(Sp) |
| Habit | Havyasa(Ka), Habit(En) Abhyasa(Sa)=Practice |
| Bull | Vrishabh(Sa) Basava(Ka) |
| Think | Yochane(Ka) Sochna(Hi) |
| Hand | Kara(Sa), Kai(Ka) |
| Light (not heavy) | Laghu(Sa), Hagura(Ka) |

| after | Apar(Sa) as in Aparhana(afternoon) Aparo(Ta) |
|---|---|
| North-East | Ishanya(Sa) Here East and Isha whose ruler is Sun God. The ruler of Ishana is Shiva |
| Eat | Tran(Sa) Root Tinnu(Ka) Tindi Tran(Sa)=Grass, Tan(Ko) = Grass,food for cattle |
| Lame | KunTa (Ka) |
| Vaikunta(Sa) | Without being Kunta(Sa): Dull, Blunt (Blunt edge to the foot?) |
| Turning around | Ulta(Hi), Volte face(En) : Turning around, changing stance |
| Steal | Steal(En) Asteya(Sa): Not stealable. Same root St |
| School | Shāla(Sa)=Branch originally then school, Shole(Ge) School(En) Ecole(Fr) Shaale(Ka) |
| Pillar | Sthambha(Sa), Khamba(Hi,Ko) Tamh(Pu) Tun(Ta) Kanua(Si), Stob(Ru) Saila)No) Syun(Ar) |
| Lake/Pond | TāTaka(Sa) TaLe(ko) Sara(Sa) Sarasu(Te) Tādakam(Ma)<br>Kere(Ka) Eri(Ta) |
| Well | Vaapi (Sa) Well(En) Bavi(Ka) |
|  | Kona(S): Angle KoNe(Ka)-Room (generally at the corner) |
| Or | Va(Sa) or(En) Ve(Ko) |
| Time | Hour(En), Hora(Sa), Hottu(Ka) |

| English | Sanskrit | Tamil | Kannada | Malayalam | Telugu |
|---|---|---|---|---|---|
| Turmeric | Haridra | Manjai | Arisina | Manyal | Pasupu |

Semantic Drift - Now to today

| Ee Hottu(Ka): This time | Ivattu(Today) |
|---|---|
| Adya(Sa) Now Adhuna(Sa) Now | Aaj(Hi), Aaaji(Ko) Today Aatt(Ko): Now |
| Adya(Sa) Now | Udya(Mar) _Tomorrow |

Roots and Words (Ram Swarup)

| Root/Basic word | Word |
|---|---|
| bhan(Sa): speak | Phone: voice or sound |
| pas(Sa)/spas(Sa):see | Telescope |
| Graphein (Gr): scratch,write | Telegraph, carve (En) |

| Tāra(Sa):star, astron(Gr), nau(Sa):boat | astronaut |
|---|---|
| Peda(Gr):steering oars | Pilot (originally plying boats) |
| daa(Sa): give | dāna(Sa), donation(En), dose, condone |
| spand(Sa): vibrate | pendere(La), pensive, pendulum, append, suspend, expend,poise, ponder |
| sthā (Sa) | sthala(place).sthāna(place), sthāa(receptacle), sthāpatya(architecture) sthapati(architect), sthira(stationary), sthitha(standing firm), sthuna(post,pillar), stand, state, stationary, statue, status, stable, sthambha (pillar) |
| bhu(sa) phynai(gr): to be born | Bhava(existence), bhavana(abode, mansion), bhuta(has been), bhavishya(future). physics, physical, buan(ge):dwell, be, build, bower, fui(La): I have been |
| Jnā (Sa):to know | Jnāna(knowledge),<br>ājnā permission), gigoskein(Ge), gnoscere(La), know, knowledge, acknowledge, gnostic, ignorant, znate(Ru), noble, cunning, keen, can, narrate |
| path | path(Sa):Path pathya:suitable food (for journey) |
| brh(Sa): To tear<br>vrasc(Sa): To cut down | Vraksha(Sa):Tree |
| rad(Sa): to bite, scratch<br>rodere(La):to gnaw | rat |
| mus(Sa,Gr,La):To steal<br>muis(Ru)<br>maus(Ge) | mushaka (Sa): mouse, mouse(En), muscle:looks like mouse |
| undare(La)<br>und(Sa), ud(Sa): to flow, bathe | Udan(water), hudor(Gr), Wanduo(Li), wasser(Ge), water(En), udāk(Ko), unda(La):wave, undulating, redundant, abundant, abound |

| vira(Sa):brave | virtue |
|---|---|
| nad(Sa):To make roaring sound | nadi(Sa):river |
| agni(Sa):Fire, ignis(La) | ugnis(Li), ogni(Sl), ignition |
| apa(Sa):water | aqua(La) |
| vāri(Sa):water | urine |
| shvān(Sa):dog | canine(La), sobaka(Ru) |
| tam(Sa):gasp of breath<br>timere(La): to fear | timid |
| vah(Sa): to move<br>uhere(La): to carry, transport | Heavy, weighty, |
| vāk(Sa):speech | Voice, vocal, vowel, vouch, invoke, evoke, revoke, provoke, advocate, vocation, convocation, equivocal, vocabulary |
| bhid(Sa):break, divide | biting |
| kuta(Sa):knife | cutting |
| svad(Sa):To taste, to eat | Sweet, hedus (Gr) |
| vid (Sa): To know, To see | eidenai( Gr):To know, idein (Gr):To see,<br>Wizze, wisdom, vidya<br>Vision, view, vista, visit<br>Veda, Vedas (known as well as seen by seers) |

## Annexure 5: Miscellaneous Words

The first two letters in the name of a language are used to indicate the language.

| Word | Word | Word/Remarks |
|---|---|---|
| tan(sa): body | trunk(en) | |
| par(hi) but | pan(ma) | |
| kadime(ka), kammi(ka) | kam(hi)<br>kami(hi) | |
| Itara(sa) | other(En) | |
| pot(ko), pet(hi) | hotte(ka) | kadupu(ta), vayiru(ta), varayu(ma) |
| Ashwaroodha(sa) | Riding horse | rooda(sa) riding(en) |
| Relief(en) | Riyayati(ka) Riyayati(Hi) | Rihai(hi)-release |
| Paleo(en):old | Hale(ka):old, palaiya(Ta):old | |
| Pilla(te):child | Pila(ko): little one of animal, child of dog or cat for example. Here sound I is longer. | Pillai(malayalam): child of King |
| Bounce, Go away. | Uchal(Hi) | UsaL(ko, ma) |
| Fire | Urja(sa) | Ujjo(ko) |
| Air | Pavan(sa) | Hava(Hi) |
| Full, too much | Tumbaa(Ka) | Ramba(Ta) |
| Gana(sa): Singing | Canadh(ir):sing | |
| Bake, cook | Pec, pecyot(ru) | Pac(sa) |
| bark | Lay-lact(ru) | Ray-rayati(sa) |
| To exist | Be-budet(ru) | Bhu-bhavati(sa) |
| To heat, hit | Tuz-tuzit(ru) | Tuj-tojayati(sa) |
| Burn, shine | Gor-gorit(ru) | Ghr-gharati(sa) |
| To caress, fondle, comfort | Las-lasket(ru) | Las-lasati(sa) |
| To cart, transport,carry, draw | Voz-vozit(ru) | Vah-vahati(sa) |
| To catch | Lov,lovit(ru) | Labh,labhate(sa):To take, seize,catch |
| To coddle, pamper, cherish, foster | Lel-lelyeet(ru) | Lal-laalayati(sa) |
| Continue to do, linger on | Bav-bavit(ru) | Bhu-bhavayati(sa): exist, be found |
| To fall | Pad-Padyot(ru) | Pad-Padyate(sa) |
| To fart | Perd-Perdit(ru) | Pard-pardati(sa) |
| To fear, be afraid | Boya-Boitsya(ru) | Bhyas-bhyasate(sa) |
| To give away | Otda-Otdayot(ru) | Udda-uddadati(sa) |
| To give out, to distribute | Vid-vidayot(ru) | Vida-vidadati(sa) |
| To give to drink | Po-poit(ru) | Pa-Payayti(sa): To give to drink to horses and Camel |
| To go, walk | i-idyot(ru) | It-etati(sa) |
| To happen, to be present, to frequent | Biv-bivaet(ru) | Bhu-bhavati(sa)- To happen, occur |
| To knead | Mes-mesit(ru) | Misr-misrayati(sa) – To mix,mingle |
| To know | Zna-znaet(ru) | Jna-Janati(sa) To know, have knowledge |
| To lick | Liz-lizet(ru) | Lih-lihati(sa) |

| aja(sa): goat | aadu(Ka) | Meke is alternate word |
|---|---|---|
| Mesha(sa) sheep | Meke(ka):goat | |
| Gariasi(sa) | Greater(en) | Jannai Janmabhoomisha Svargadapi Gariyasi. |
| Varaaha(sa) | Boar(en) | |
| Ede(ka)- Chest | Hradaya(sa)- heart | Harde(ko)-Chest |
| Chest(en) | Chhati(hi)-Chest | |
| Vekh(pu):See | View(en),<br>Dekh(hi),<br>Vision(en) | Gaze, See |
| Vastra(sa) | Batte(ka) | |
| Mad(en): Mad | Mada(sa):Arrogant, Haughty, Lost its composure | |
| Taayi(ka):Mother<br>Taata(ka):Grand Father | Taata(sa):Father | Taayi and Taata go together as mother and father. |
| Sthula(sa) Fat Wide | Storas(li)<br>Thora(ko) – Fat<br>Tali? (ru) | |
| Maha(sa)- great | Mahato(_) -Fat<br>Mota (hi) -Fat | |
| Shankha(sa) – doubt, suspicion | Shak(h) – Doubt, suspicion. | |